\documentclass[10pt,journal,compsoc]{IEEEtran}
\ifCLASSOPTIONcompsoc
  \usepackage[nocompress]{cite}
\else
  \usepackage{cite}
\fi
\usepackage[hyphens]{url}
\usepackage{hyperref}
\hypersetup{colorlinks=true,breaklinks=true}
\usepackage{dsfont}
\usepackage{amsmath}
\usepackage{float}
\usepackage{mathtools}
\usepackage[T1]{fontenc}
\usepackage{amsmath}
\usepackage{caption}
\usepackage{subcaption}
\usepackage{textalpha}
\usepackage{dsfont}
\usepackage{amsmath}
\usepackage{float}
\usepackage{multirow}
\usepackage{hhline}
\usepackage{longtable}
\usepackage{caption}
\usepackage{xcolor,pifont}
\usepackage{multicol}
\usepackage[left=17.5mm,right=17.5mm,top=24.5mm,bottom=33.95mm]{geometry}
\usepackage{tabularx}
\usepackage{amssymb}
\usepackage{pifont}
\usepackage{adjustbox}
\usepackage{lipsum}
\usepackage{booktabs}
\usepackage{xcolor}

\usepackage{xcolor,pifont}
\newcommand*\colourcheck[1]{%
  \expandafter\newcommand\csname #1check\endcsname{\textcolor{#1}{\ding{52}}}%
}
\colourcheck{blue}
\colourcheck{green}
\colourcheck{red}
\usepackage{multirow}
\usepackage{hhline}
\usepackage{times}
\usepackage{epsfig}
\usepackage{graphicx}
\usepackage{amsmath}
\usepackage{amssymb}
\usepackage{colortbl}
\usepackage[hyphens]{url}
\usepackage{hyperref}
\hypersetup{colorlinks=true,breaklinks=true}
\usepackage{dsfont}
\usepackage{amsmath}
\usepackage{float}
\usepackage{mathtools}
\usepackage[T1]{fontenc}
\usepackage{amsmath}
\usepackage{caption}
\usepackage{subcaption}
\usepackage{textalpha}
\usepackage{dsfont}
\usepackage{amsmath}
\usepackage{float}
\usepackage{multirow}
\usepackage{hhline}
\usepackage{longtable}
\usepackage{caption}
\usepackage{xcolor,pifont}
\usepackage{multicol}
\usepackage[left=17.5mm,right=17.5mm,top=24.5mm,bottom=33.95mm]{geometry}
\usepackage{tabularx}
\usepackage[utf8]{inputenc}
\usepackage{graphicx}
\usepackage{wrapfig}

\usepackage{tikz}
\usepackage{forest}
\usetikzlibrary{trees,positioning,shapes,shadows,arrows.meta}
\usepackage{amssymb}
\usepackage{pifont}
\usepackage{booktabs}

\ifCLASSINFOpdf
\else
\fi
\begin{document}
%
\title{Semi-Supervised Object Detection: A Survey on
Progress from CNN to Transformer}

\author{Tahira~Shehzadi,
        Ifza,
        Didier~Stricker
        and~Muhammad~Zeshan~Afzal
\IEEEcompsocitemizethanks{\IEEEcompsocthanksitem All the members are with Department of Computer Science Technical University of Kaiserslautern, Mindgarage Lab, 
German Research Institute for Artificial Intelligence (DFKI)
Kaiserslautern, Germany 67663 \protect\\
E-mail: Tahira.Shehzadi@dfki.de 
}
}
\IEEEtitleabstractindextext{%
\begin{abstract}
The impressive advancements in semi-supervised learning have driven researchers to explore its potential in object detection tasks within the field of computer vision. Semi-Supervised Object Detection (SSOD) leverages a combination of a small labeled dataset and a larger, unlabeled dataset. This approach effectively reduces the dependence on large labeled datasets, which are often expensive and time-consuming to obtain. Initially, SSOD models encountered challenges in effectively leveraging unlabeled data and managing noise in generated pseudo-labels for unlabeled data. However, numerous recent advancements have addressed these issues, resulting in substantial improvements in SSOD performance. This paper presents a comprehensive review of 27 cutting-edge developments in SSOD methodologies, from Convolutional Neural Networks (CNNs) to Transformers. We delve into the core components of semi-supervised learning and its integration into object detection frameworks, covering data augmentation techniques, pseudo-labeling strategies, consistency regularization, and adversarial training methods. Furthermore, we conduct a comparative analysis of various SSOD models, evaluating their performance and architectural differences. We aim to ignite further research interest in overcoming existing challenges and exploring new directions in semi-supervised learning for object detection.

\end{abstract}

}

\maketitle

\IEEEdisplaynontitleabstractindextext

%
\IEEEpeerreviewmaketitle

\IEEEraisesectionheading{\section{Introduction}\label{sec:introduction}}
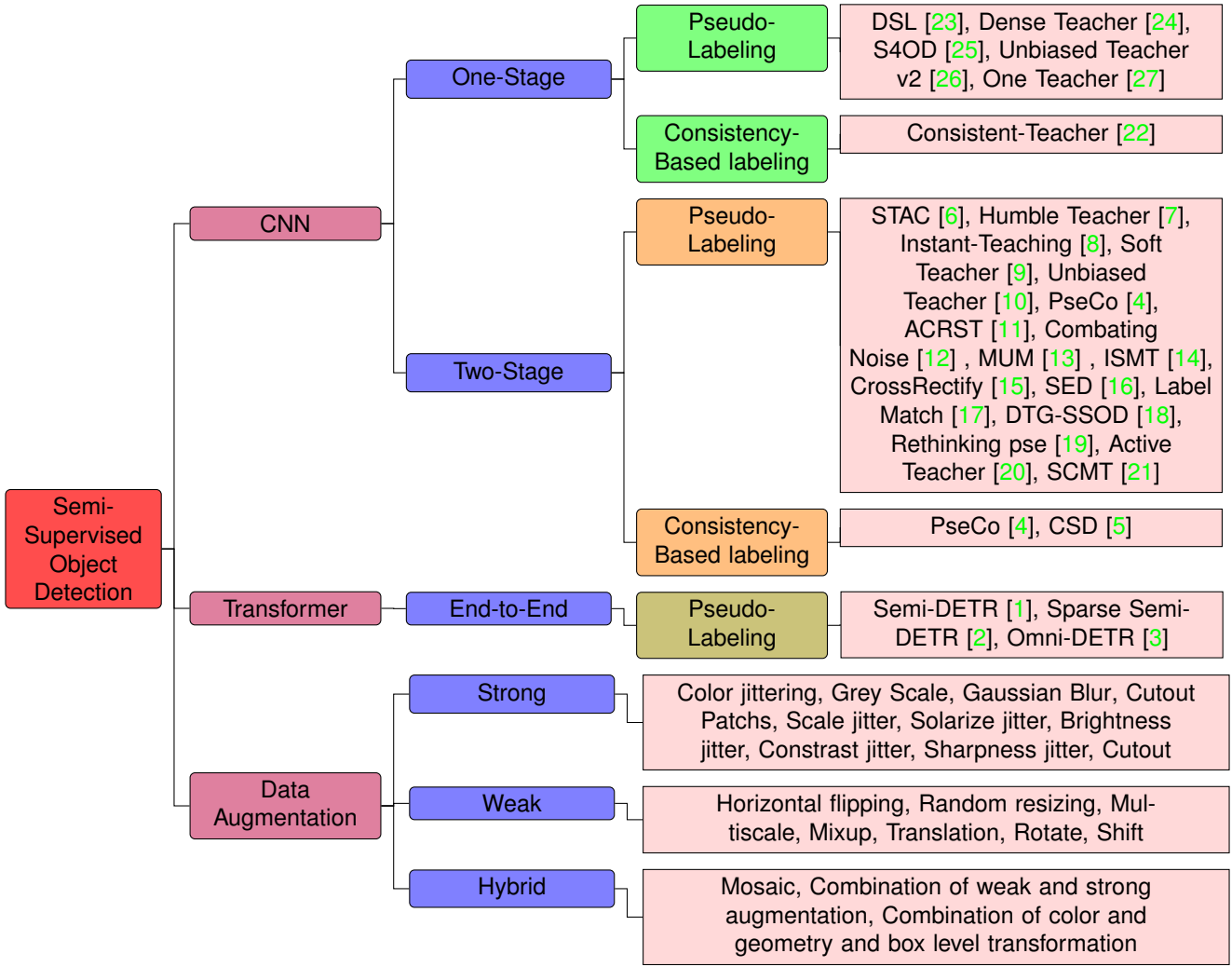
\begin{figure*}
\centering
\tikzset{
    basic/.style  = {draw, text width=2cm, align=center, fill=red!60, font=\sffamily, rectangle},
    root/.style   = {basic, rounded corners=2pt, thin, align=center, fill=green!30},
    onode/.style = {basic, thin, rounded corners=2pt, align=center, fill=pink!60,text width=3.5cm,},
    tnode/.style = {basic, thin, align=left, fill=pink!60, text width=15em, align=center},
    tknode/.style = {basic, thin, align=left, fill=pink!60, text width=15.6em, align=center},
    ttnode/.style = {basic, thin, align=left, fill=pink!60, text width=24.8em, align=center},
    tt2node/.style = {basic, thin, align=left, fill=pink!60, text width=24.3em, align=center},
    xnode/.style = {basic, thin, rounded corners=2pt, align=center, fill=blue!50,text width=2.7cm},
    gnode/.style = {basic, thin, rounded corners=2pt, align=center, fill=green!50,text width=2.5cm},
    ornode/.style = {basic, thin, rounded corners=2pt, align=center, fill=orange!50,text width=2.5cm},
    olnode/.style = {basic, thin, rounded corners=2pt, align=center, fill=olive!50,text width=2.5cm},
    prnode/.style = {basic, thin, rounded corners=2pt, align=center, fill=purple!50,text width=2.5cm},
    cynode/.style = {basic, thin, rounded corners=2pt, align=center, fill=cyan!50,text width=2.5cm},
    tlnode/.style = {basic, thin, rounded corners=2pt, align=center, fill=teal!50,text width=2.5cm},
    snode/.style = {basic, thin, rounded corners=2pt, align=center, fill=red!70,text width=2cm,},
    wnode/.style = {basic, thin, align=left, fill=pink!10!blue!80!red!10, text width=6.5em},
    edge from parent/.style={draw=black, edge from parent fork center}
}
\centering
\begin{forest}
for tree={
    grow=east,
    parent anchor=east,
    child anchor=west,
    edge path={\noexpand\path[draw, \forestoption{edge}]
             (!u.parent anchor) -- +(5pt,0pt) |-  (.child anchor)
             \forestoption{edge label};
        }
}
[Semi-Supervised Object Detection, snode
    [Data Augmentation, prnode
        [Hybrid, xnode
            [Mosaic{,} Combination of weak and strong augmentation{,} Combination of color and geometry and box level transformation, tt2node]
        ]
        [Weak, xnode
            [Horizontal flipping{,} Random resizing{,} Multiscale{,} Mixup{,} Translation{,} Rotate{,} Shift, tt2node]
        ]
        [Strong, xnode
            [Color jittering{,}  Grey Scale{,} Gaussian Blur{,} Cutout Patchs{,}  Scale jitter{,}  Solarize jitter{,}  Brightness jitter{,}  Constrast jitter{,} Sharpness jitter{,}  Cutout, tt2node]
        ]
    ]
     [Transformer, prnode,  l sep=3.5mm,
        [End-to-End, xnode, l sep=3.5mm,
        [Pseudo-Labeling, olnode, l sep=1.8mm,
         [Semi-DETR~\cite{Semi-DETR_cvpr23}{,} Sparse Semi-DETR~\cite{sparse_semi_detr2}{,} Omni-DETR~\cite{omni-DETR_cvpr22}, tknode]]
         ]
         ]
     [CNN, prnode,  l sep=3.5mm,
       [Two-Stage, xnode,  l sep=3.5mm,
        [Consistency-Based labeling, ornode, l sep=1.7mm,
         [PseCo~\cite{Pseco99}{,} CSD~\cite{CSD75}, tknode]]
        [Pseudo-Labeling, ornode, l sep=1.7mm,
         [STAC~\cite{STAC54}{,} Humble Teacher~\cite{HumbleTeacher56}{,} Instant-Teaching~\cite{InstantTeaching59}{,} Soft Teacher~\cite{SoftTeacher}{,} Unbiased Teacher~\cite{UnbiasedTeacher57}{,} PseCo~\cite{Pseco99}{,} ACRST~\cite{ACRST73}{,} Combating Noise~\cite{Combating92} {,} MUM~\cite{Mum95} {,} ISMT~\cite{ISMT94}{,} CrossRectify~\cite{Rectify93}{,} SED~\cite{SED97}{,} Label Match~\cite{Label94}{,} DTG-SSOD~\cite{DTG95}{,} Rethinking pse~\cite{Rethinking98}{,} Active Teacher~\cite{Active100}{,} SCMT~\cite{SCMT22} ,tknode]] 
    ]     
    [One-Stage, xnode,  l sep=3.5mm,
        [Consistency-Based labeling, gnode,  l sep=1.7mm,
          [Consistent-Teacher~\cite{Consistent-Teacher72},tknode]]
        [Pseudo-Labeling, gnode, l sep=1.7mm
         [DSL~\cite{DSL_CVPR22}{,} Dense Teacher~\cite{Dense-Teacher74}{,} S4OD~\cite{S4OD70}{,} Unbiased Teacher v2~\cite{UnbiasedTeacherv2_CVPR22}{,} One Teacher~\cite{OneT96},tknode]]
     ]
    ]
    ]
\end{forest}

\caption{Semi-Supervised Object Detection: A Comprehensive Review and Taxonomy of Techniques.}
\label{fig:lit_surv}
\end{figure*}
\begin{table*}[h!]
\tiny
\begin{center}
\renewcommand{\arraystretch}{0.7} 
\begin{tabular*}{\textwidth}{@{\extracolsep{\fill}}p{6.6cm}p{0.5cm}p{9.8cm}@{\extracolsep{\fill}}}
\toprule
\textbf{Title} & 
\textbf{Year} &
\textbf{Description}  \\
\toprule

Semi-Supervised Learning Literature Survey~\cite{Survey85} & 2008  & This survey examines the landscape of semi-supervised learning literature concentrating on diverse methodologies and applications.
\\
\midrule
A Survey On Semi-Supervised Learning Techniques
~\cite{Survey78} & 2014 & An Analysis investigates various techniques in semi-supervised learning, offering insights into their effectiveness and applications.
\\
\midrule
A Survey and Comparative Study of Tweet Sentiment Analysis via Semi-Supervised Learning~\cite{Survey90} & 2016   & 
This study provides a thorough comparison and analysis of tweet sentiment methods employing semi-supervised learning techniques.
\\
\midrule
Semi-supervised learning for medical application: A survey~\cite{Survey79} & 2018  & This paper delves into the integration of semi-supervised learning within medical contexts, offering insights into its applicability and potential advancements.
  \\
\midrule
A survey on semi supervised learning\cite{Survey82} & 2019  & This comprehensive examination explores the domain of semi-supervised learning, shedding light on its practical implementations and advancements.\\
\midrule
Improvability Through Semi-Supervised Learning: A Survey of Theoretical Results~\cite{Survey87} & 2020  & This analysis investigates theoretical advancements facilitated by semi-supervised learning, exploring avenues for improvement within machine learning frameworks.
\\
\midrule
Small Data Challenges in Big Data Era: A
Survey of Recent Progress on Unsupervised
and Semi-Supervised Methods~\cite{Survey88} & 2020   & This exploration examines recent progress in unsupervised and semi-supervised methods, addressing challenges posed by small data in the context of the big data era.
\\
\midrule

A Survey of Un-, Weakly-, and Semi-Supervised Learning Methods for Noisy, Missing and Partial Labels in Industrial Vision Applications
~\cite{Survey91} & 2021   & This survey evaluates unsupervised, weakly-supervised, and semi-supervised learning techniques designed to address problems caused by noisy, incomplete, and missing labels in industrial vision applications.
\\
\midrule
Semi-Supervised and Unsupervised Deep Visual Learning: A Survey
~\cite{Survey86} & 2022   & This study explores the field of deep visual learning, with a particular focus on semi-supervised and unsupervised methods. It aims to uncover key insights and advancements in these approaches.
\\
\midrule
A Survey on Semi-, Self- and Unsupervised Learning for Image Classification~\cite{Survey89} & 2022   & This survey examines image classification, focusing on semi-supervised, self-supervised, and unsupervised learning methods to understand their effectiveness and potential applications.
\\
\midrule
A survey on semi-supervised learning for delayed partially labelled data streams \cite{Survey83}& 2022  & This study delves into semi-supervised learning approaches employed for handling delayed data streams with semi labels, focusing on their effectiveness and challenges.\\
\midrule
Semi Supervised deep learning for image classification with distribution mismatch: A survey\cite{Survey84}& 2022  & This study explores Semi-Supervised Deep Learning for image classification with distribution mismatch, providing insights into its strategies and challenges. \\
\midrule
A Survey on Deep Semi-supervised Learning~\cite{Survey76} & 2023   & This survey examines the field of deep semi-supervised learning techniques, providing insights into their applications and advancements.
\\
\midrule
Graph-based semi-supervised learning: A comprehensive review~\cite{Survey80}  & 2023  & This comprehensive review examines the effectiveness and applications of graph-based semi-supervised learning methods. \\
\bottomrule
\end{tabular*}
\caption{Overview of previous surveys on object detection. For each paper, the publication details are provided.}\label{tab:survey}

\end{center}
\end{table*} 
Deep learning~\cite{intro_deep,intro_deep2,intro_deep3,intro14} has become an active area of research with numerous applications in various fields such as pattern recognition~\cite{intro_pattern, intro_pattern2}, data mining~\cite{intro_mining,intro_mining3}, statistical learning~\cite{intro_statistical,intro_statistical2}, computer vision~\cite{intro_vision1, intro_vision2}, and natural language processing~\cite{intro_NLP1,intro_NLP2,intro_NLP1}. It has seen significant achievements particularly in supervised learning contexts, by effectively utilizing a substantial amount of high-quality labeled data. However,
these supervised learning approaches~\cite{intro7,introsupervised,introsupervised1}, rely on labeled data for training that is costly and time-consuming. Semi-Supervised Object Detection (SSOD)~\cite{intro5} bridges this gap by incorporating both labeled and unlabeled data~\cite{inrosemisupervised}. It shows a significant advancement in the field of computer vision~\cite{intro_vision1,intro_vision2}, particularly for industries where obtaining extensive labeled data~\cite{intro5} is challenging or costly. SSOD is used in various sectors, including Autonomous vehicles~\cite{app_3d3,app_3d4} as well as medical imaging~\cite{app_image1,app_image2}. In industries like agriculture~\cite{introagri1}~\cite{introagri2} and manufacturing~\cite{intro_manu1}, where there's lots of data but labeling is time-consuming, SSOD helps make things more efficient.

Semi-supervised methods~\cite{intro_semimethods,intro_semimethods2} enhance model performance and reduce labeling needs by employing both unlabeled and labeled data~\cite{intro_model,Intro_model2}. Moreover, previous object detection~\cite{intro1,intro2} approaches primarily involved manual feature engineering~\cite{intro_feature1,intro_feature2} and the use of simplistic models. These approaches faced difficulties in accurately identifying objects with different shapes and dimensions~\cite{intro3}. Later, the introduction of Convolutional Neural Networks (CNNs)~\cite{intro4,intro_CCN} revolutionizes object detection by directly extracting hierarchical features~\cite{intro_CCN3} from raw data, enabling end-to-end learning~\cite{intro_CNN2} and substantially enhancing accuracy and effectiveness. In recent years, Semi-Supervised Object Detection has made significant improvement, driven by advancements in deep learning architectures~\cite{intr_arch1,intro_arch2}, optimization techniques~\cite{intro_opt1}, and dataset augmentation strategies~\cite{intro_aug1,introdataaug,Stac_augmen1,Instant_augmen1}. Researchers have developed various semi-supervised learning (SSL) approaches tailored for object detection, each with distinct strengths and limitations~\cite{intro10,intro_survey1,intro_survey2}. These approaches are mainly categorized into pseudo-labeling~\cite{intropseudo,intropseudo2,intropseudo3} and consistency regularization~\cite{intro9}, both of which effectively utilize labeled and unlabeled data during training. Moreover, the integration of SSL methods with state-of-the-art object detection architectures such as FCOS~\cite{FCOS}, Faster R-CNN~\cite{faster101}, and YOLO~\cite{yolo45} has significantly enhanced the performance and scalability of Semi-Supervised Object Detection systems. This combination not only improves detection accuracy but also helps models work well with new and unseen datasets.

Object detection has seen remarkable progress with the advent of the DEtection TRansformer(DETR)~\cite{detr1,detr2,shehzadi2023object5}. Transformers, originally developed for natural language processing~\cite{intro_NLP1,intro_NLP2,intro_NLP1}, excel in capturing long-range dependencies~\cite{intro_dependencies} and contextual information~\cite{intro_contextual,intro_contextual1}, making them ideal for complex spatial arrangements~\cite{intro_spatail, intro_spatail1} in object detection. Unlike CNNs~\cite{intro_CCN,intro_CNN2,intro_CCN3}, which rely on localized convolutions and require non-maximum suppression (NMS)~\cite{intro_NMS} to filter out redundant detections, DETR uses self-attention mechanisms~\cite{intro_attent1, intro_attent2} and do not need NMS. It considers the object detection task as a direct set prediction problem, eliminating traditional processes like NMS~\cite{intro_NMS} and anchor generation~\cite{intro_anchor}. Despite its advantages, DETR has limitations, such as slow convergence during training and challenges with small object detection. To address these issues, advancements in DETR enhance performance and efficiency through improved attention mechanisms and optimization techniques~\cite{intro_optdetr}. Following DETR's success, researchers are now employing DETR-based networks in Semi-Supervised Object Detection approaches\cite{Semi-DETR_cvpr23,sparse_semi_detr2,omni-DETR_cvpr22}. This combines DETR's strengths with semi-supervised learning to use unlabeled data\cite{intro9,intro10}, reducing the need for large labeled datasets. 

Due to the rapid progress of transformer-based Semi-Supervised Object Detection (SSOD)\cite{inrosemisupervised, inrosemisupervised2} approaches, keeping up with the latest advancements has become increasingly challenging. Therefore, a review of ongoing developments from CNN-based to Transformer-based SSOD methods is essential and would greatly benefit researchers in the field. This paper presents a comprehensive overview of the transition from CNN-based to Transformer-based approaches in Semi-Supervised Object Detection (SSOD). As shown in Fig.~\ref{fig:lit_surv}, the survey categorizes SSOD approaches into CNN-based (one-stage and two-stage)~\cite{intro14,Dense-Teacher74,UnbiasedTeacher57,OneT96,STAC54,HumbleTeacher56,InstantTeaching59,Pseco99} and Transformer-based approaches~\cite{Semi-DETR_cvpr23,sparse_semi_detr2,omni-DETR_cvpr22}, highlighting techniques like pseudo-labeling and consistency-based labeling. It also provide details about data augmentation strategies~\cite{introdataaug,Stac_augmen1,Stac_augmen2,unbiased_augment1,unbiased_augment2,Instant_augmen2,Instant_augmen1}, including strong, weak, and hybrid techniques.\par
\begin{figure}[h]
\centering
\includegraphics[width=\linewidth]{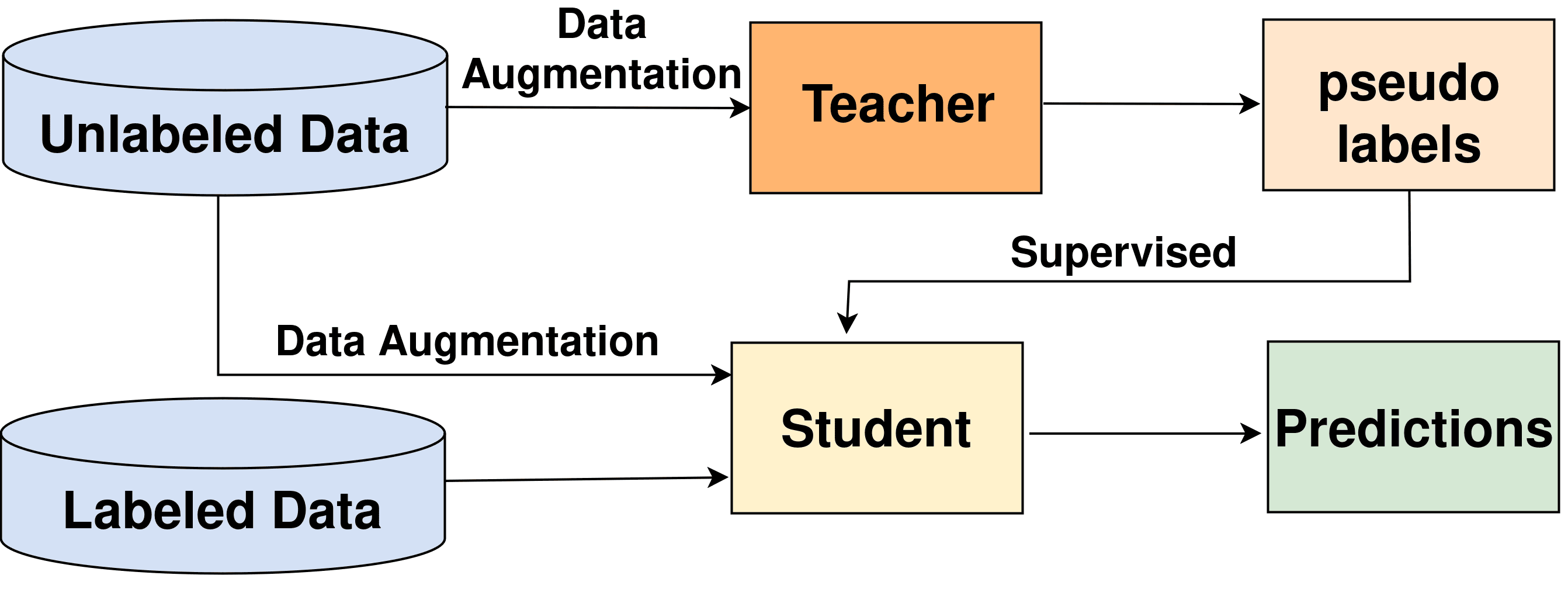}
\caption{Teacher-Student Architecture for Semi-Supervised Object Detection}\label{fig:general_fig}
\end{figure}

Fig.\ref{fig:general_fig} depicts a teacher-student architecture tailored for semi-supervised object detection. A pretrained teacher model is utilized to generate pseudo-labels for unlabeled data. These pseudo-labels, along with the labeled data, are then utilized to jointly train the student model. By incorporating pseudo-labeled data, the student model learns from a more extensive and diverse dataset, enhancing its ability to detect objects accurately. Additionally, data augmentation methods are applied to both labeled and pseudo-labeled datasets. This collaborative learning approach effectively leverages both labeled and unlabeled data to improve the overall performance of object detection systems.

The remainder of this paper is organized as follows: Section~\ref{sec:surveys} provides a review of previous surveys on SSOD. Section~\ref{sec:related_work} discusses related work in the field. Section~\ref{sec:application} explores the role of SSODs in various vision tasks. Section~\ref{sec:all_semi}, the core of this paper, offers a comprehensive overview of SSOD approaches. Section~\ref{sec:loss} examines different loss functions used in SSOD. Section~\ref{sec:comparison} presents a comparative analysis of SSOD approaches. Section~\ref{sec:Discussion} addresses open challenges and future directions. Finally, Section~\ref{sec:conclusion} concludes the paper.

\section{RELATED PREVIOUS REVIEWS AND SURVEYS}
\label{sec:surveys}
Table~\ref{tab:survey} provides an overview of previous surveys on object detection, highlighting key research in semi-supervised learning. It covers a range of topics from theoretical advancements\cite{Survey85,Survey87} to practical applications\cite{Survey91} across various domains. These surveys investigate diverse methodologies and their effectiveness, including specific applications in tweet sentiment analysis~\cite{Survey90} and medical contexts~\cite{Survey78}. Recent works explore improvements within machine learning frameworks~\cite{Survey82}, addressing challenges posed by small data and industrial applications with noisy or incomplete labels~\cite{Survey91}. Notably, some surveys focus on deep visual learning and image classification using semi-supervised~\cite{Survey77,Survey88}, self-supervised~\cite{Survey78,Survey89}, and unsupervised methods~\cite{Survey86}, providing valuable insights into their effectiveness and challenges. Collectively, these surveys offer a detailed understanding of the advancements, challenges, and practical implementations in the field of Semi-Supervised Object Detection. While previous surveys have focused on CNN-based SSOD methods, the rise of Transformer-based Semi-Supervised Object Detection requires thorough evaluation to understand their effectiveness and trends.

\section{Related Work}
\label{sec:related_work}
Semi-Supervised Object Detection (SSOD)~\cite{inrosemisupervised,inrosemisupervised2,intro10,inrosemisupervised3,inrosemisupervised4} has achieved remarkable progress with diverse approaches that utilize both labeled data and generate labels for unlabeled data to improve model performance. This section provides an overview of key contributions and methodologies in SSOD.
\subsection{Early Semi-Supervised Approaches}
Early approaches in semi-supervised learning for object detection aimed to adapt self-training techniques~\cite{RW_self,RW_self2} from image classification~\cite{RW_selftoclass,RW_selftoclass2} to object detection. STAC~\cite{STAC54} (Self-Training with Consistency), which employs a two-stage process: first, it generates high-confidence pseudo-labels~\cite{STAC_pseudo} from unlabeled images; then, it trains the model using both labeled and pseudo-labeled data with strong augmentations~\cite{Stac_augmen1,Stac_augmen2} to ensure consistency~\cite{Stac_Consistency1,Stac_Consistency2}.Another example is Unbiased Teacher~\cite{UnbiasedTeacher57}, which uses a teacher-student framework where the teacher model generates pseudo-labels for the student model. The student model is then trained with these pseudo-labels~\cite{intropseudo,intropseudo2,intropseudo3} along with labeled data, while the teacher model is updated with exponential moving averages of the student’s weights to improve stability and robustness.
\subsection{Teacher-Student Frameworks}
Recent advancements have introduced sophisticated teacher-student frameworks~\cite{RW_studteach,RW_studteach2,semimask4} that include additional mechanisms to enhance SSOD. In these frameworks, a teacher model generates pseudo-labels~\cite{intropseudo,intropseudo2,intropseudo3} from unlabeled data, which are then used to train a student model, thereby improving the student’s performance iteratively. For example, the Consistent-Teacher framework~\cite{Consistent-Teacher72} seeks to minimize inconsistent pseudo-targets through adaptive anchor assignment~\cite{Consist_ASA1,Consist_ASA2}, feature alignment~\cite{Consist_tood}. Similarly, Dense Teacher guidance frameworks ~\cite{Dense-Teacher74} improve the quality of pseudo-labels~\cite{intropseudo,intropseudo2,intropseudo3} by utilizing dense predictions from the teacher model.

\subsection{Consistency Regularization}
Consistency regularization~\cite{Stac_regular1, Stac_Consistency1,Stac_Consistency2} in SSOD ensures that the model generates consistent predictions across different augmented views of the same image, promoting robustness and generalization. The Mean Teacher~\cite{Mean-Teacher71} framework, utilizing a teacher-student paradigm, has been adapted for Semi-Supervised Object Detection (SSOD). Techniques such as Interactive Self-Training with Mean Teachers~\cite{ISMT94} build on this approach by iteratively refining pseudo-labels~\cite{intropseudo,intropseudo2,intropseudo3} and enhancing the performance of the student model. Additionally, employing consistency regularization~\cite{Stac_regular1, Stac_Consistency1,Stac_Consistency2}, where models are trained to generate consistent predictions under varying augmentations, has proven effective in boosting SSOD performance.

\subsection{Pseudo-Labeling Methods}
Pseudo-labeling~\cite{intropseudo,intropseudo2,intropseudo3}, which involves the model generating labels for unlabeled data, is another fundamental aspect of SSOD. Techniques like Rethinking Pseudo~\cite{Rethinking98} Labels introduce improvements to conventional pseudo-labeling~\cite{intropseudo2,intropseudo3} by tackling challenges such as label noise and confidence thresholds. Additionally, label matching~\cite{Label94} and dense pseudo-labeling refine this process further, ensuring the generated labels are more precise and dependable.

\subsection{Self-Training}
Self-training~\cite{RW_self,RW_self2} in semi-supervised involves iteratively generating pseudo-labels~\cite{intropseudo,intropseudo2,intropseudo3} for unlabeled data and integrating them into the training process alongside labeled data, aiding model improvement over iterations. Enhancements to self-training frameworks\cite{RW_self,RW_self2}, such as incorporating active learning strategies where the model actively selects the most informative samples for labeling, have shown promise. The Active Teacher framework\cite{Active100} is an example where the teacher model guides the selection of samples that are likely to improve the student model's learning.

\subsection{Transformer-Based Approaches}
Transformer-based approaches~\cite{Semi-DETR_cvpr23,omni-DETR_cvpr22,sparse_semi_detr2} utilize transformer architectures\cite{intotransformer,introCNNtransformer,introCNNtransformer2}, known for their ability to capture long-range dependencies, to improve detection performance by effectively modeling spatial relationships and contextual information within visual data.With the rise of transformer-based architectures\cite{intotransformer,introCNNtransformer,introCNNtransformer2}, researchers have begun integrating these models into SSOD. Semi-DETR~\cite{Semi-DETR_cvpr23}, for example, adapts the Detection Transformer(DETR)~\cite{detr1,detr2,shehzadi2023object5} model to a semi-supervised setting, demonstrating the potential of transformers in improving detection performance in SSOD tasks.

\section{Semi Supervised Strategies}
\label{sec:all_semi}

\subsection{One Teacher} 
\begin{figure}[H]
\centering
\includegraphics[width=0.80\linewidth]{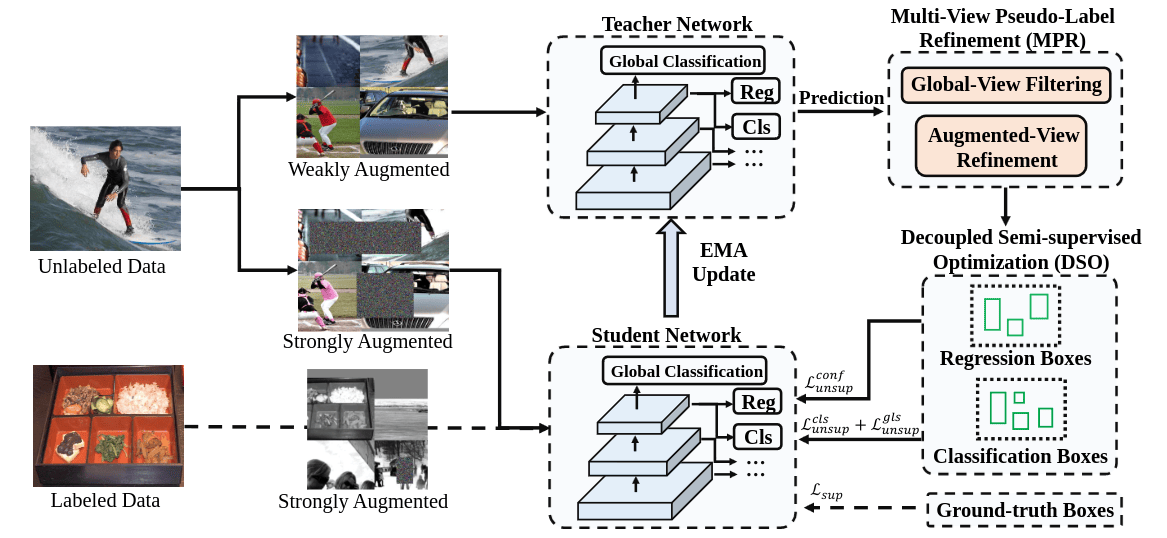}
   \caption{ \textbf{Framework of One Teacher~\cite{OneT96}} }
\label{fig:One Teacher}
\vspace{-10pt}
\end{figure}

With a focus on the advanced Yolov5 model~\cite{OneT_YOLO5,OneT_YOLO}, One Teacher~\cite{OneT96} presents a novel teacher-student learning strategy designed especially for one stage Semi-Supervised Object Detection (SSOD), as illustrated in Fig. \ref{fig:One Teacher}. By addressing the fundamental issues of one-stage SSOD, such as inefficient pseudo-labeling~\cite{intropseudo,intropseudo2,intropseudo3} and conflicts in multi-task optimization~\cite{OneTmultitask}, One Teacher aims to close this gap. OneTeacher optimizes teacher-student learning for one-stage SSOD using creative techniques like Multi-view Pseudo-label Refinement (MPR)~\cite{OneTmultiview} and Decoupled Semi-supervised Optimization(DSO)~\cite{OneTOpt}.
\subsection{DSL} 
\begin{figure}[H]
\centering
\includegraphics[width=0.99\linewidth]{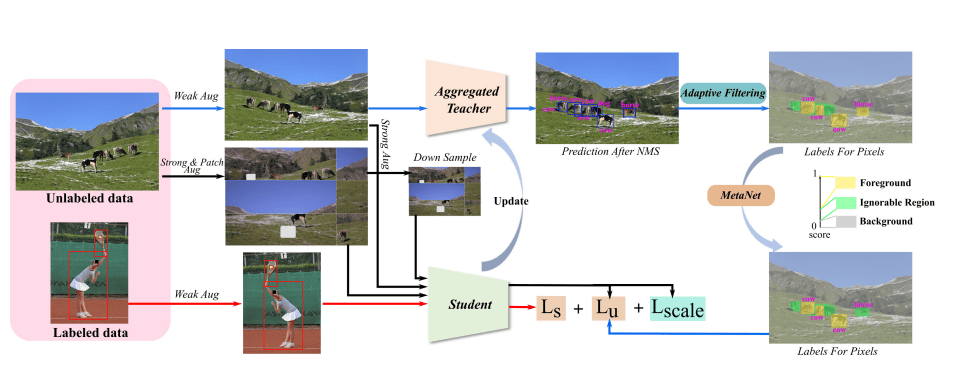}
   \caption{ \textbf{Framework of DSL~\cite{DSL_CVPR_22}} }
\label{fig:DSL}
\vspace{-10pt}
\end{figure}

DenSe Learning (DSL)~\cite{DSL_CVPR22} algorithm presents a approach to anchor-free SSOD. As shown in Fig. \ref{fig:DSL}, is designed for one stage anchor-free detector like FCOS~\cite{DSL_Fcos}, in contrast to current approaches that mainly concentrate on two stage anchor-based detectors, which are more practical for real-world applications. DSL addresses key challenges by introducing innovative techniques such as Adaptive Filtering (AF) for precise pseudo-label assignment~\cite{Rethinking98, adaptivefilter}, Aggregated Teacher (AT)~\cite{aggregate1} for enhanced label stability, and uncertainty consistency regularization~\cite{Consistency_SED} for improved model generalization.
\subsection{Dense Teacher} 
\begin{figure}[H]
\centering
\includegraphics[width=0.80\linewidth]{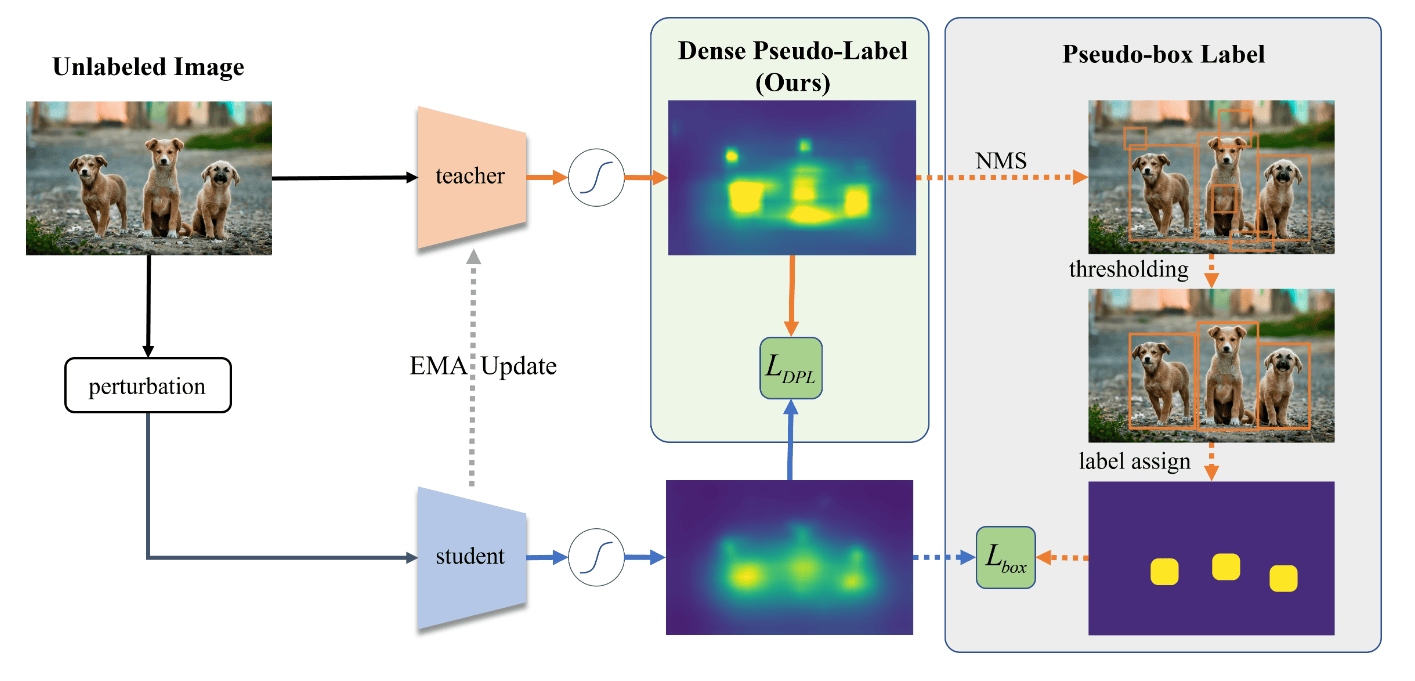}
   \caption{ \textbf{Framework of Dense Teacher~\cite{DenseTeacher}} }
\label{fig:Dense Teacher}
\vspace{-10pt}
\end{figure}

The Dense Teacher~\cite{Dense-Teacher74} framework introduces a innovative approach to Semi-Supervised Object Detection (SSOD) by replacing sparse pseudo-boxes with dense predictions termed Dense Pseudo-Labels(DPL)~\cite{DLP2,DLP1}, as demonstrated in Fig. \ref{fig:Dense Teacher}.
Post-processing procedures, such as Non-Maximum Suppression~\cite{intro_NMS}, are not necessary for this unified pseudo-label~\cite{regression_Plabels,intropseudo,intropseudo2} structure. Additionally, a region division strategy is proposed to suppress noise and enhance the focus on key regions, further improving detection accuracy. Overall, Dense Teacher represents a significant advancement in SSOD with its streamlined pipeline and effective utilization of dense pseudo-labels~\cite{DLP2,DLP1}.

\subsection{Unbiased Teacher v2} 
\begin{figure}[H]
\centering
\includegraphics[width=0.80\linewidth]{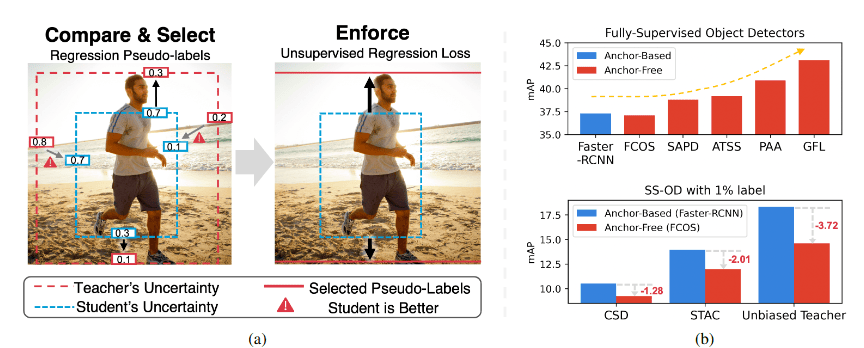}
   \caption{ \textbf{Framework of Dense Teacher~\cite{DenseTeacher}} }
\label{fig:Unbiased Teacher v2}
\vspace{-10pt}
\end{figure}

Unbiased Teacher v2~\cite{UnbiasedTeacherv2_CVPR22} introduces an innovative method that extends the scope of SSOD techniques~\cite{STAC54, SoftTeacher58,HumbleTeacher56,Rectify93} to anchor-free detectors, alongside the introduction of the Listen2Student mechanism to unsupervised regression loss~\cite{STAC54,InstantTeaching59} is depicted in Fig. \ref{fig:Unbiased Teacher v2}.Key contributions include expanding the applicability of SSOD to both anchor-based and anchor-free detectors~\cite{Anchorfree_based}, developing a mechanism to address misleading instances in regression pseudo-labels~\cite{regression_Plabels,intropseudo,intropseudo2}, and reducing performance differences between anchor-free and anchor-based detectors~\cite{Anchorfree_based} in the semi-supervised domain.
\subsection{S4OD} 
\begin{figure}[H]
\centering
\includegraphics[width=0.90\linewidth]{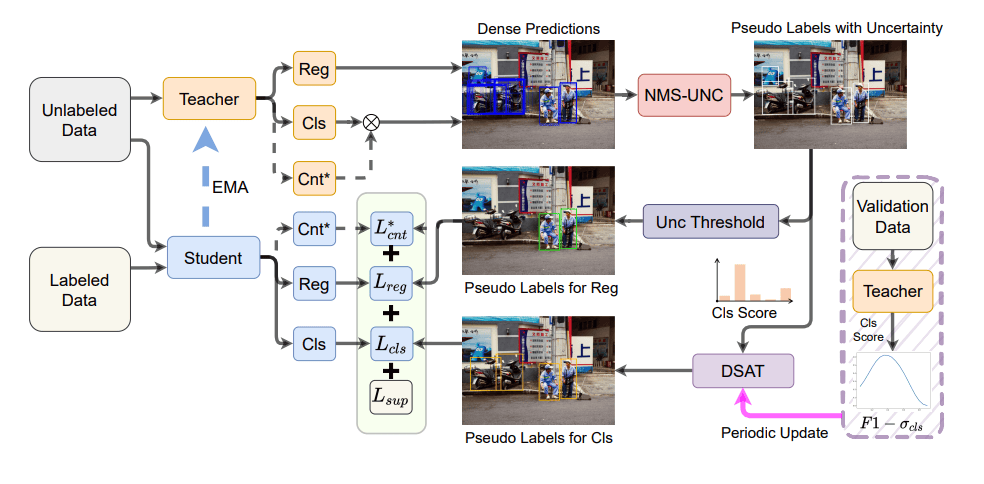}
   \caption{ \textbf{Framework of S4OD~\cite{S4OD70}} }
\label{fig:S4OD}
\vspace{-10pt}
\end{figure}

S4OD~\cite{S4OD70}, a semi-supervised methodology tailored for one stage detectors, addresses the challenge of extreme class imbalance~\cite{S4OD_Imblanace} inherent in these detectors compared to their two stage SSOD~\cite{STAC54, SoftTeacher58,HumbleTeacher56}. Shown in Fig. \ref{fig:S4OD}, S4OD introduces the Dynamic Self-Adaptive Threshold (DSAT) strategy~\cite{DSAT}. S4OD dynamically determines pseudo-label selection~\cite{intropseudo, intropseudo2, intropseudo3}, balancing label quality and quantity in the classification branch. Additionally, the NMS-UNC module evaluates regression label quality by computing box uncertainties via Non-Maximum Suppression~\cite{intro_NMS}, enhancing regression targets.~\cite{SoftTeacher,unbiased_Teacher_ICLR21}
\subsection{Consistent-Teacher} 
Inconsistent pseudo labels~\cite{intropseudo, intropseudo2,intropseudo3} in Semi-Supervised Object Detection (SSOD) pose a challenge that Consistent-Teacher~\cite{Consistent-Teacher72} addresses. These pseudo labels introduce noise into the student's training process, which causes serious overfitting~\cite{overfitting1} problems and compromises the construction of accurate detectors.  
\begin{figure}[H]
\centering
\includegraphics[width=0.99\linewidth]{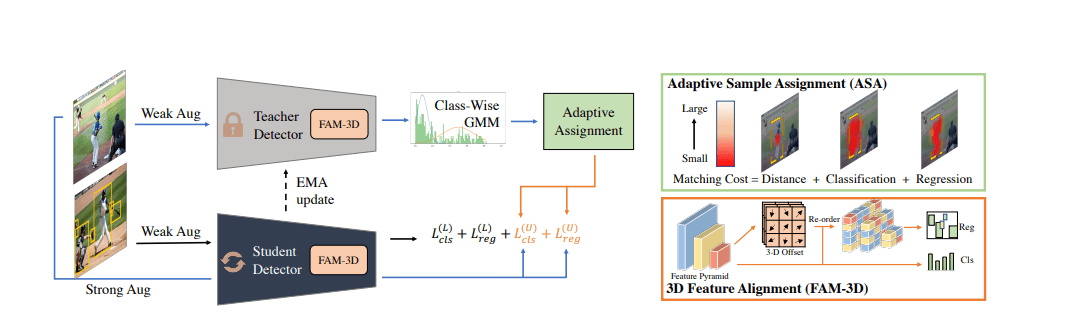}
   \caption{ \textbf{Framework of Consistent-Teacher~\cite{Consistent-Teacher72}} }
\label{fig:Consistent Teacher}
\vspace{-10pt}
\end{figure}
As represented in Fig. \ref{fig:Consistent Teacher}, Consistent-Teacher introduces a 3D feature alignment module (FAM- 3D)~\cite{Consist_tood}, the Gaussian Mixture Model (GMM), and adaptive anchor assignment (ASA)~\cite{Consist_ASA1,Consist_ASA2} as a strategy to minimize this issue.  These components enhance the quality of the pseudo-boxes, dynamically modify the threshold values, and stabilize the pseudo-box matching with anchors.

\subsection{Rethinking Pse} 
\begin{figure}[H]
\centering
\includegraphics[width=0.95\linewidth]{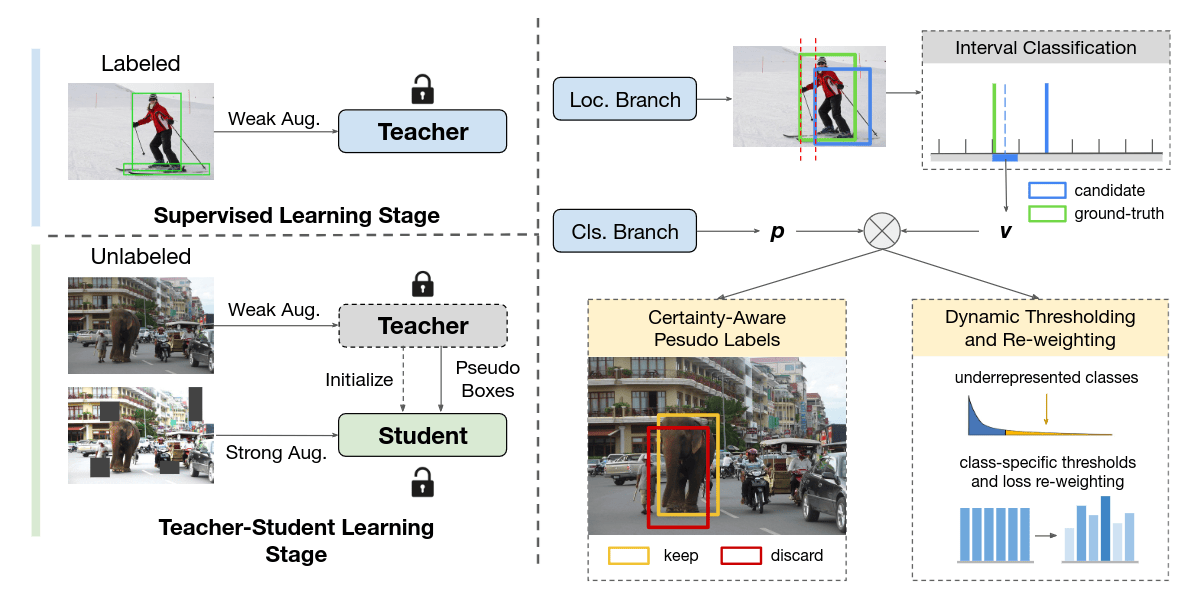}
   \caption{ \textbf{Framework of Rethinking Pse~\cite{Rethinking98}} }
\label{fig: Rethinking Pse}
\vspace{-10pt}
\end{figure}

Rethinking Pse ~\cite{Rethinking98}, as shown in Fig.\ref{fig: Rethinking Pse}, introduces certainty aware pseudo labels~\cite{intropseudo, intropseudo2,intropseudo3} that are specifically designed for object detection. These labels accurately assess the quality of both classification and localization~\cite{pseco_localization_2}, providing a more refined method for generating pseudo labels~\cite{intropseudo, intropseudo2,intropseudo3},. By dynamically adjusting thresholds and reweighting loss functions~\cite{lossfun} based on these certainty measurements, this mitigates the challenges posed by class imbalance~\cite{imbalance_ACRST1,imbalance_ACRST2,imbalance_ACRST3,imbalance_ACRST4,S4OD_Imblanace}.

\subsection{CSD} 
\begin{figure}[H]
\centering
\includegraphics[width=0.95\linewidth]{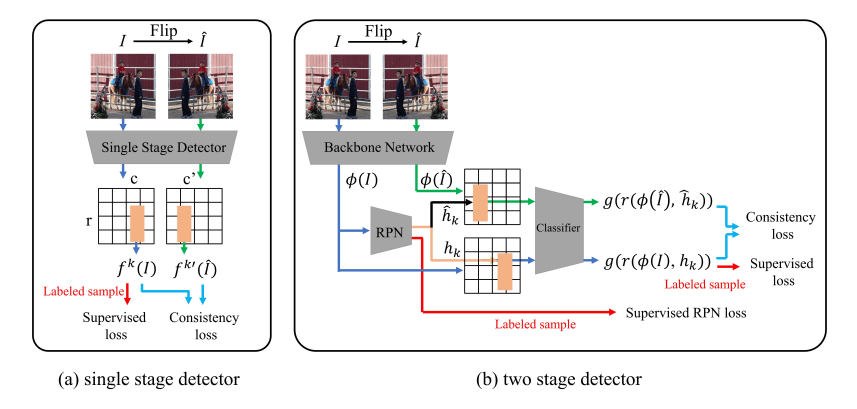}
   \caption{ \textbf{Framework of CSD~\cite{CSD75}} }
\label{fig:CSD}
\vspace{-10pt}
\end{figure}

CSD~\cite{CSD75}(Consistency-based Semi-supervised learning method for object Detection), which utilizes consistency constraints~\cite{consistCons} to maximize the use of accessible unlabeled data and improve detection performance, as illustrated in Fig. \ref{fig:CSD}. This approach extends beyond object classification to include localization~\cite{pseco_localization_2}, ensuring comprehensive model training~\cite{pseco_localization_2}. Additionally, this introduces Background Elimination(BE) to lessen the adverse effects of background noise on detection accuracy. 

\subsection{STAC}
\begin{figure}[H]
\centering
\includegraphics[width=0.95\linewidth]{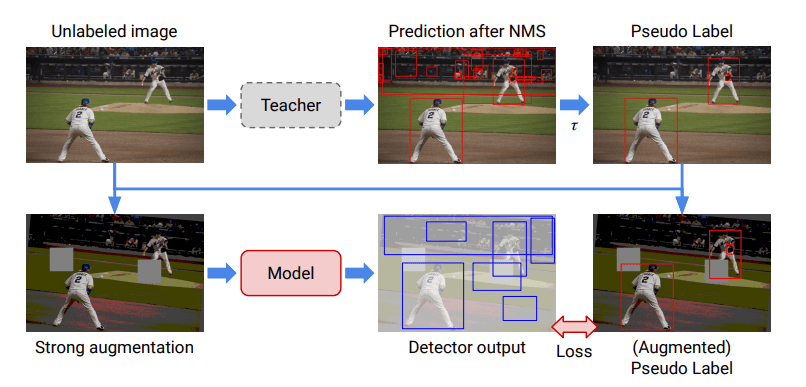}
\caption{ \textbf{Overview of STAC~\cite{STAC54}} }
\label{fig:STAC}
\vspace{-10pt}
\end{figure}

STAC~\cite{STAC54} is a semi-supervised~\cite{inrosemisupervised,inrosemisupervised2} framework designed to enhance detection models for visual object recognition using unlabeled data, as shown in Fig. \ref{fig:STAC}. The baseline detector employed in the proposed architecture is Faster R-CNN~\cite{faster101}. It follows a two-step procedure where a trained detector is utilized in the first stage to generate high-confidence pseudo-labels~\cite{STAC_pseudo} from unlabeled images. To ensure consistency and robustness, the model undergoes further training in the second stage using labeled and pseudo-labeled data along with significant data augmentations~\cite{Stac_augmen1,Stac_augmen2}. STAC combines augmentation-driven consistency regularization~\cite{Stac_regular1} and self-training~\cite{Stac_Consistency1,Stac_Consistency2} to extend the state-of-the-art SSL from image classification~\cite{Survey77}~\cite{Survey88} to object detection.
\subsection{Humble Teacher}

\begin{figure}[H]
\centering
\includegraphics[width=0.99\linewidth]{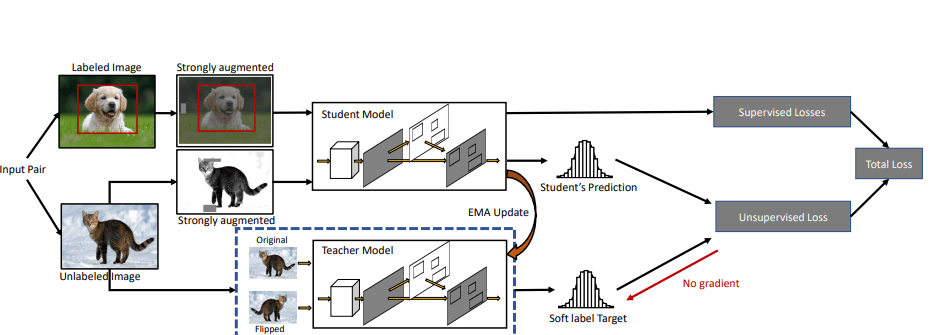}
   \caption{ \textbf{Overview of Humble Teacher~\cite{HumbleTeacher56}} }
\label{fig:Humble Teacher}
\vspace{-10pt}
\end{figure}

Humble Teacher~\cite{HumbleTeacher56} proposes semi-supervised approach for contemporary object detectors, utilizing a teacher-student dual model framework,as illustrated in Fig. \ref{fig:Humble Teacher}. The method incorporates dynamic updates to the teacher model through exponential moving averaging (EMA)~\cite{confirmationbias_meanT}, employs soft pseudo-labels and multiple region proposals as training targets for the student, and utilizes a detection-specific data ensemble to generate more dependable pseudo-labels. Unlike existing approaches such as STAC~\cite{STAC54}, which rely on hard labels for sparsely selected pseudo samples, the method leverages soft-labels on multiple proposals, allowing the student to distill richer information from the teacher~\cite{Humblerichinfo}.

\subsection{Instant-Teaching} 
\begin{figure}[H]
\centering
\includegraphics[width=0.99\linewidth]{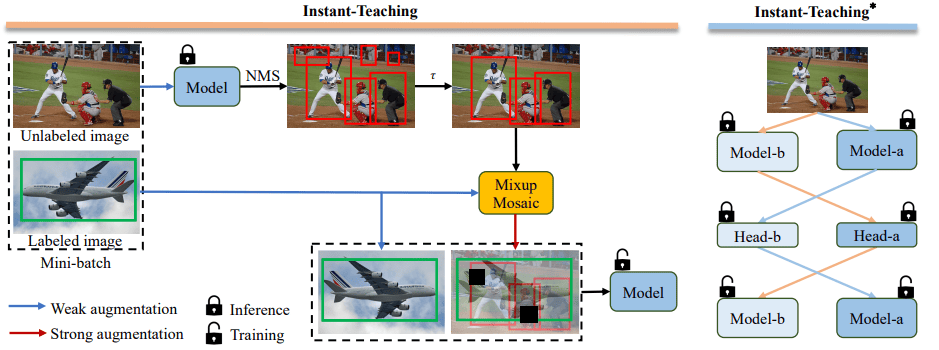}
   \caption{ \textbf{Overview of Instant Teaching~\cite{InstantTeaching59}} }
\label{fig:Instant Teaching}
\vspace{-10pt}
\end{figure}

Instant-Teaching\cite{InstantTeaching59} leverages instant pseudo labeling~\cite{intropseudo, intropseudo2,intropseudo3} and extended weak-strong data augmentations~\cite{Instant_augmen1,Instant_augmen2,Instant_augmen3}~\cite{Instant_augmen4} throughout each training iteration to overcome the limitations of manual annotations in typical supervised object detection frameworks. The system implements Instant-Teaching, a co-rectify approach~\cite{Rectify93}, to improve pseudo annotation quality and reduce confirmation bias~\cite{confirmationbias_meanT}, as depicted in Fig. \ref{fig:Instant Teaching}.

\subsection{ISMT} 
\begin{figure}[H]
\centering
\includegraphics[width=0.90\linewidth]{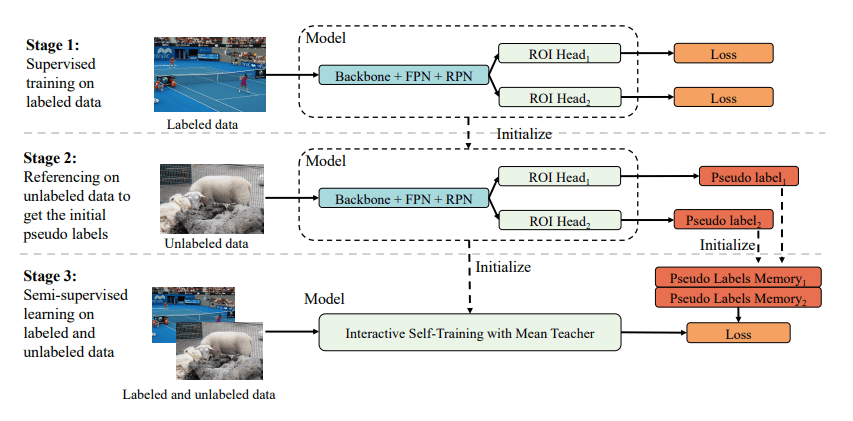}
   \caption{ \textbf{Framework of ISMT~\cite{ISTM55}} }
\label{fig:ISTM}
\vspace{-10pt}
\end{figure}

A Semi-Supervised Object Detection technique known as Interactive Self-Training with Mean Teachers (ISMT)~\cite{ISTM55} introduces an approach to rectify the oversight of inconsistencies among detection outcomes in the same image across various training iterations, as shown in Fig. \ref{fig:ISTM}. By utilizing non maximum suppression~\cite{intro_NMS} to combine detection outcomes from different iterations and employing multiple detection heads to offer complementary information, this approach boosts the stability and quality of pseudo labels. Moreover, the incorporation of the mean teacher model~\cite{Mean-Teacher71} prevents overfitting~\cite{overfitting1} and aids in the transfer of knowledge between detection heads.

\subsection{Combating Noise} 
\begin{figure}[H]
\centering
\includegraphics[width=1.00\linewidth]{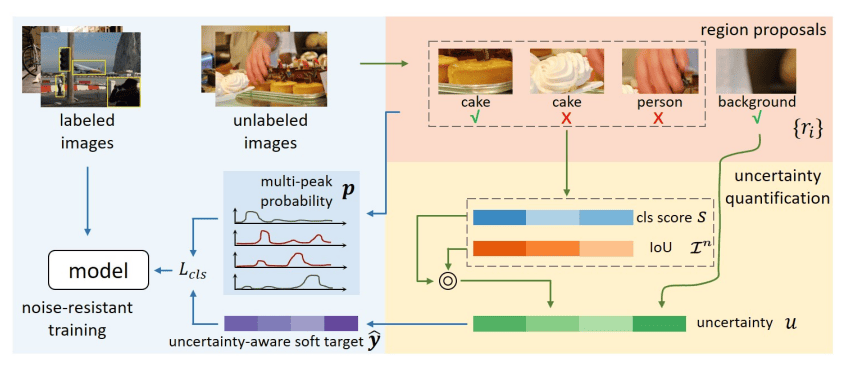}
   \caption{ \textbf{Framework of Combating Noise~\cite{Combating92}} }
\label{fig:Combating Noise}
\vspace{-10pt}
\end{figure}

The proposal outlined in Combating Noise~\cite{Combating92} introduces a method resilient to noise by measuring region uncertainty to mitigate the negative impacts of noisy pseudo-labels~\cite{NoisyPL1,Noisy_Pseudo}. With this method, the effects of noisy pseudo-labels are carefully examined, and a metric for measuring region uncertainty is ultimately developed. By incorporating this metric into the learning framework~\cite{metric_learning},an uncertainty-aware soft target can be formulated to prevent performance degradation caused by noisy pseudo-labeling~\cite{NoisyPL1}, as illustrated in Fig. \ref{fig:Combating Noise}. Additionally, it mitigates overfitting~\cite{overfitting1} by allowing multi-peak probability distributions and removing competition among classes.
\subsection{Soft Teacher} 
\begin{figure}[H]
\centering
\includegraphics[width=0.85\linewidth]{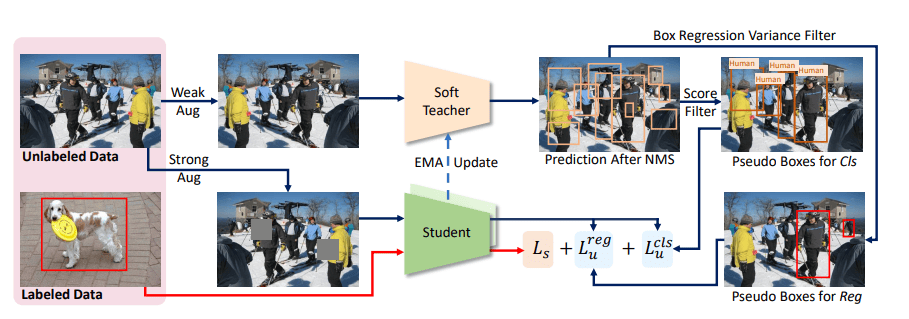}
   \caption{ \textbf{Overview of Soft Teacher~\cite{SoftTeacher58}} }
\label{fig:Soft Teacher}
\vspace{-10pt}
\end{figure}

In contrast to earlier multi-stage approaches, Soft Teacher~\cite{SoftTeacher58} introduces an end-to-end solution for Semi-Supervised Object Detection. The object detection training efficiency is increased by this new framework, which progressively enhances pseudo label~\cite{intropseudo, intropseudo2,intropseudo3} attributes during training~\cite{softpre,STAC54}. As shown in Fig. \ref{fig:Soft Teacher}, this framework proposes two straightforward yet efficient methods: a box jittering methodology~\cite{box_jittering_1} for choosing robust pseudo boxes for box regression learning~\cite{Bound_box_Reg1}, and a soft teacher mechanism involving classification loss is balanced by the classification score from the teacher network.

\subsection{Unbiased Teacher} 
\begin{figure}[H]
\centering
\includegraphics[width=0.90\linewidth]{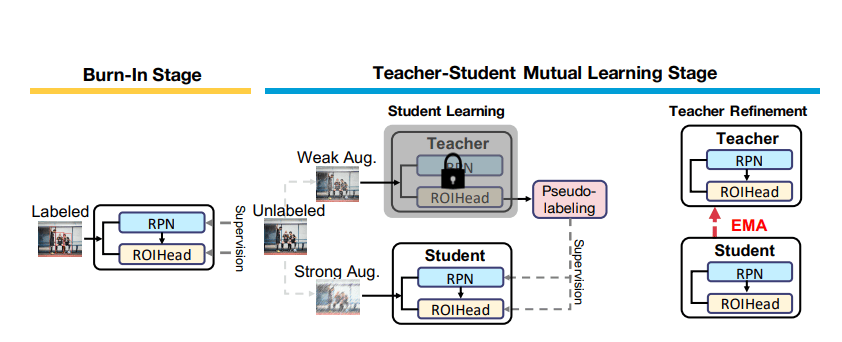}
   \caption{ \textbf{Overview of Unbiased Teacher~\cite{UnbiasedTeacher57}} }
\label{fig:Unbaised Teacher}
\vspace{-10pt}
\end{figure}

Unbiased Teacher~\cite{UnbiasedTeacher57} framework tackles the bias issue in pseudo-labeling~\cite{intropseudo, intropseudo2,intropseudo3}, prevalent in SSOD due to class imbalances~\cite{imbalance_ACRST1,imbalance_ACRST2,S4OD_Imblanace}, as shown in Fig. \ref{fig:Unbaised Teacher}. By collaborating to train a student and a teacher, who learns slowly, Unbiased Teacher leverages Exponential Moving Average (EMA)~\cite{EMA1} and differential data augmentation~\cite{unbiased_augment1,unbiased_augment2,unbiased_augment3} to enhance pseudo-label quality and mitigate overfitting~\cite{overfitting1}.
The approach addresses key challenges in SSOD, including class imbalance and overfitting, leading to notable performance enhancements in object detection.
 
\subsection{DTG-SSOD} 
\begin{figure}[H]
\centering
\includegraphics[width=0.98\linewidth]{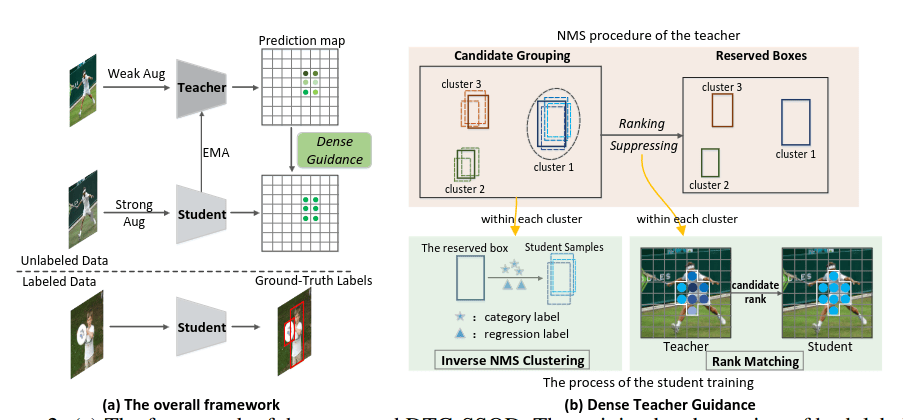}
   \caption{ \textbf{Framework of DTG-SSOD~\cite{DTG95}} }
\label{fig:DTG-SSOD}
\vspace{-10pt}
\end{figure}

Using the 'dense-to-dense' methodology, Dense teacher Guidance for Semi-Supervised Object Detection (DTG-SSOD)~\cite{DTG95} utilizes dense teacher predictions directly to guide student training. 
As represented in Fig. \ref{fig:DTG-SSOD}, this method is facilitated through techniques such as Inverse NMS Clustering (INC)and Rank Matching (RM)~\cite{DTG95}, allows the student model to emulate the teacher's behavior during Non-Maximum Suppression (NMS)~\cite{DTG_NMS}, thereby receiving dense supervision without relying on sparse pseudo labels. INC clusters candidate boxes similar to the teacher's NMS process, while RM aligns the score rank of clustered candidates between the teacher and student. 
\subsection{MUM} 
MUM\cite{Mum95}, a data augmentation approach~\cite{unbiased_augment1,unbiased_augment2,unbiased_augment3}, is introduced to tackle challenges in effectively utilizing strong data augmentation strategies in SSOD due to potential adverse effects on bounding box localization~\cite{Instant_augmen2}. 
\begin{figure}[H]
\centering
\includegraphics[width=.98\linewidth]{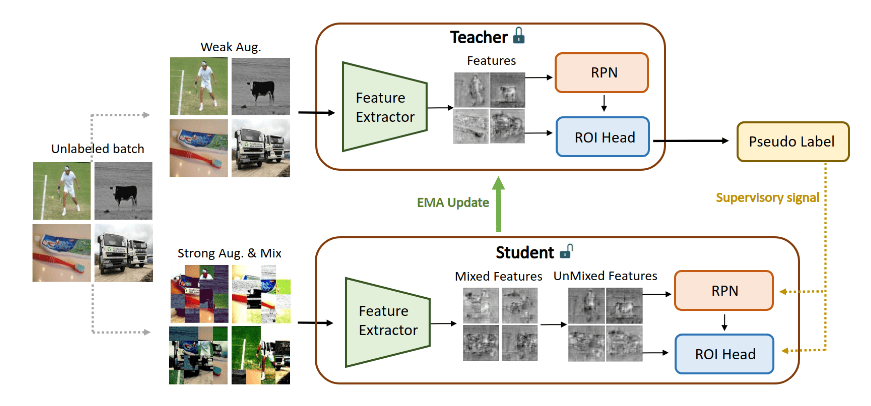}
   \caption{ \textbf{Framework of MUM~\cite{Mum95}} }
\label{fig:MUM}
\vspace{-10pt}
\end{figure} 
As depicted in Fig. \ref{fig:MUM}, MUM facilitates mixing and reconstructing feature tiles from mixed image tiles, leveraging interpolation-regularization (IR)~\cite{MUMMUMInterpolation2} for meaningful weak-strong pair generation~\cite{MUMpair1,MUMInterpolation1}.Unlike traditional SSL methods, MUM allows for the preservation of spatial information crucial for accurate object localization.
\subsection{Active Teacher} 
\begin{figure}[H]
\centering
\includegraphics[width=0.99\linewidth]{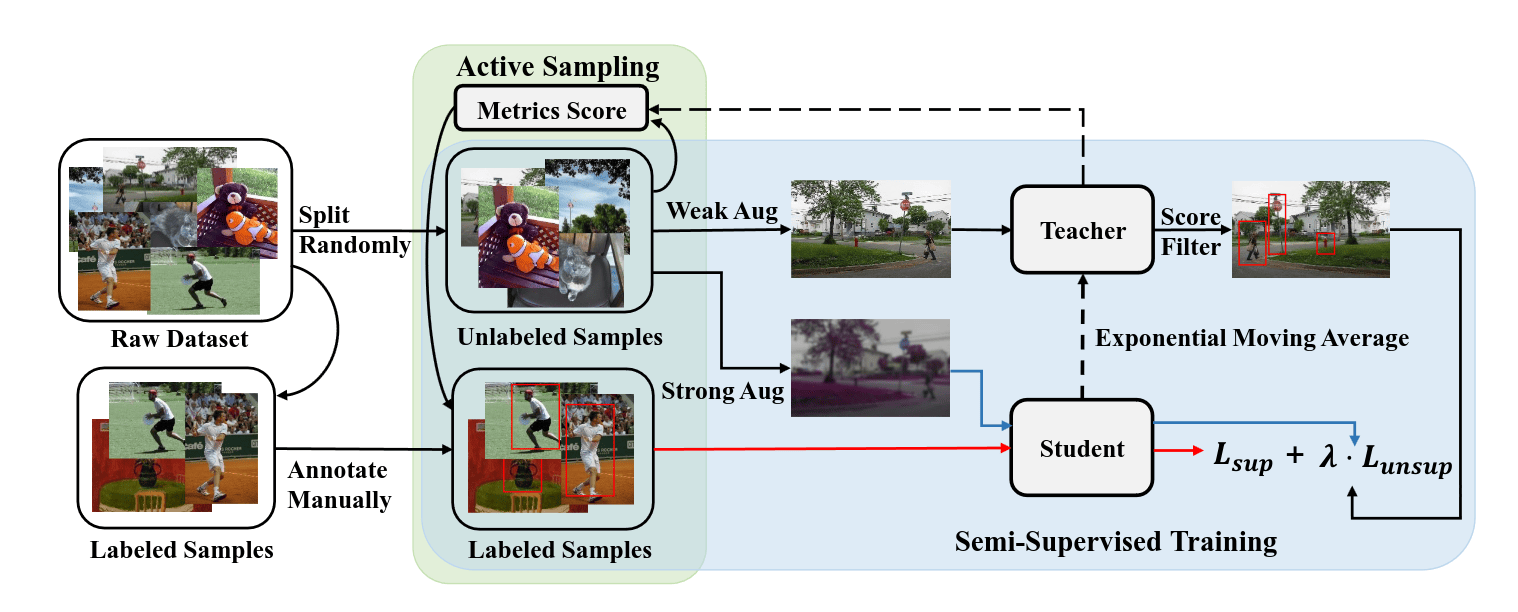}
   \caption{ \textbf{Framework of Active Teacher~\cite{Active100}} }
\label{fig:Active Teacher}
\vspace{-10pt}
\end{figure}

 Iteratively extending the teacher-student structure, the Active Teacher~\cite{Active100} method is used for Semi-Supervised Object Detection (SSOD), as demonstrated in Fig. \ref{fig:Active Teacher}. Active Teacher addresses the challenge of data initialization in SSOD by gradually augmenting~\cite{introdataaug,Stac_augmen1,Instant_augmen1} the label set through an active sampling strategy, considering factors such as difficulty, information, and diversity of unlabeled examples. Active Teacher significantly enhances the performance of SSOD by maximizing the utility of limited label information and improving the accuracy of pseudo-labels~\cite{intropseudo,intropseudo2,intropseudo3}.
\subsection{PseCo} 
Two essential strategies, pseudo-labeling and consistency training (PseCo)~\cite{Pseco99}, in Semi-Supervised Object Detection (SSOD), highlight the shortcomings of these approaches in terms of efficiently using unlabeled data for learning.
\begin{figure}[H]
\centering
\includegraphics[width=0.95\linewidth]{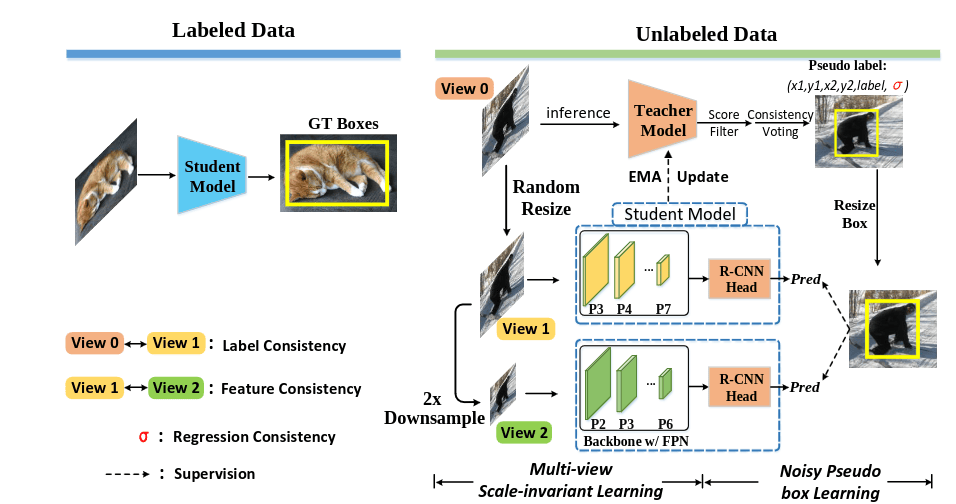}
   \caption{ \textbf{Framework of PseCo~\cite{Pseco99}} }
\label{fig:PseCo}
\vspace{-10pt}
\end{figure} 
Specifically, while existing pseudo labeling~\cite{intropseudo,intropseudo2,intropseudo3} approaches focus solely on classification scores, neglecting the precision of pseudo boxes localization,\cite{pseco_localization_1,pseco_localization_2} and commonly adopted consistency training methods overlook feature-level consistency crucial for scale invariance. To address these limitations, Noisy Pseudo box Learning (NPL)~\cite{NoisyPL1,Noisy_Pseudo} is proposed for robust pseudo label generation and Multi-view Scale-invariant Learning (MSL)~\cite{variant} is introduced to ensure both label consistency and feature-level consistency, shown in Fig. \ref{fig:PseCo}.  
\subsection{CrossRectify} 
\begin{figure}[H]
\centering
\includegraphics[width=1.05\linewidth]{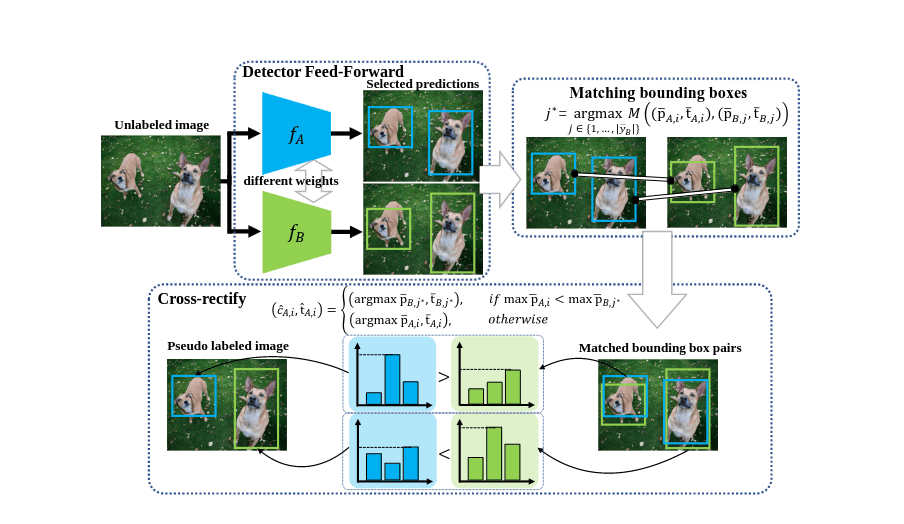}
   \caption{ \textbf{Framework of Cross Rectify~\cite{Rectify93}} }
\label{fig:Cross Rectify}
\vspace{-10pt}
\end{figure}

CrossRectify~\cite{Rectify93} is a detection framework designed to enhance the accuracy of pseudo labels~\cite{intropseudo, intropseudo2,intropseudo3}, by concurrently training two detectors with different initial parameters, as depicted in Fig. \ref{fig:Cross Rectify}. By utilizing the disparities between the detectors, CrossRectify implements a cross-rectifying mechanism~\cite{Rectify93} to identify and improve pseudo labels, thereby addressing the inherent constraints of self-labeling~\cite{self-labeling_rectify} techniques. Extensive experiments conducted across 2D~\cite{shehzadi2023object5} and 3D~\cite{3D_rectify} detection datasets validate the efficacy of CrossRectify in surpassing existing Semi-Supervised Object Detection methods.
\subsection{Label Match} 
\begin{figure}[H]
\centering
\includegraphics[width=0.95\linewidth]{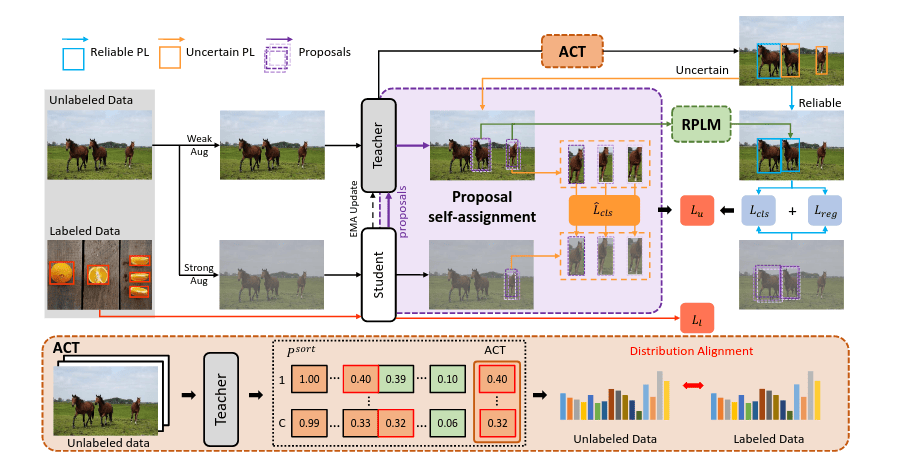}
   \caption{ \textbf{Framework of Label Match~\cite{Label94}} }
\label{fig:Label Match}
\vspace{-10pt}
\end{figure}

Label mismatch is tackled from both distribution-level and instance-level perspectives through the Label Match~\cite{Label94} architecture, shown in Fig. \ref{fig:Label Match}. A re-distribution mean teacher~\cite{Mean-Teacher71} employs adaptive label-distribution-aware~\cite{Label-Distrib} confidence criteria for unbiased pseudo-label~\cite{unbiased} creation to address distribution-level incompatibilities~\cite{SoftTeacher58,UnbiasedTeacher57,STAC}. By incorporating student suggestions into the teacher's guidance, a proposal self-assignment technique resolves instance-level mismatches stemming~\cite{stemming,stemming1} from label assignment uncertainty. Furthermore, the utilization of a reliable pseudo label mining technique~\cite{pseudo_labelmining} enhances efficiency by converting ambiguous pseudo-labels into dependable ones.

\subsection{ACRST} 
\begin{figure}[H]
\centering
\includegraphics[width=.95\linewidth]{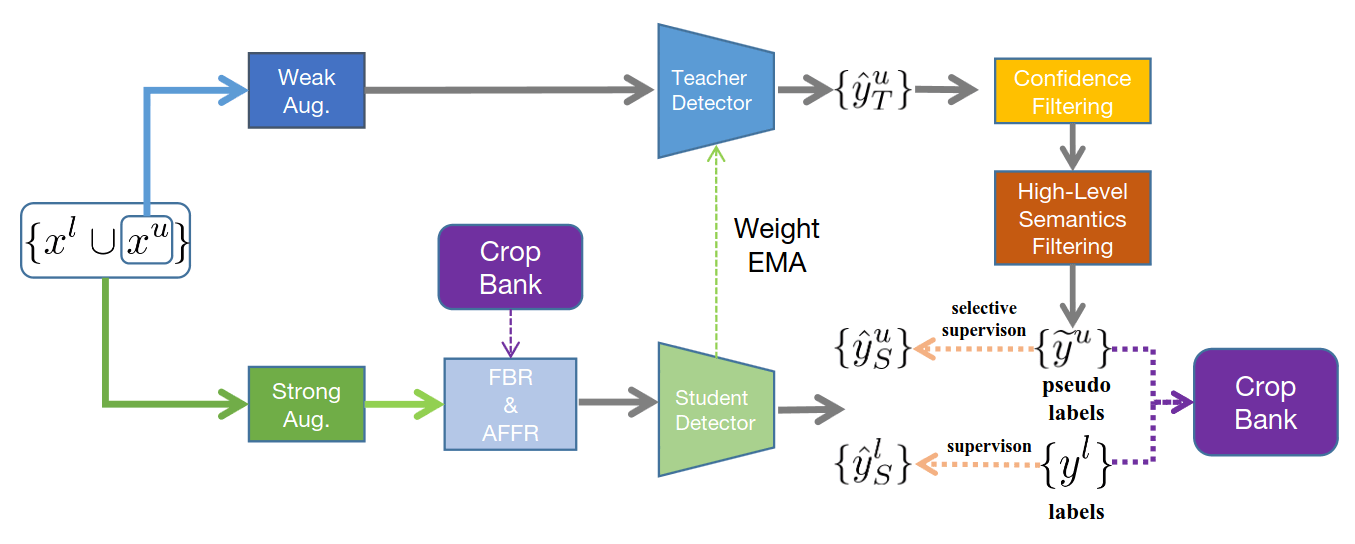}
\caption{ \textbf{Framework of ACRST~\cite{ACRST73}} }
\label{fig: ACRST}
\vspace{-10pt}
\end{figure}

Adaptive class-rebalancing self-training, or ACRST~\cite{ACRST73}, as illustrated in Fig. \ref{fig: ACRST}, introduces a new memory module called CropBank to address the major problem of class imbalance~\cite{imbalance_ACRST1,imbalance_ACRST2} in SSOD. In SSOD, class imbalance~\cite{imbalance_ACRST3,imbalance_ACRST4}, especially foreground-background and foreground-foreground imbalances—presents serious difficulties that impact the quality of pseudo-labels~\cite{intropseudo, intropseudo2,intropseudo3} and the performance of resulting models. By incorporating foreground examples from the CropBank, ACRST dynamically rebalances the training data, thereby reducing the effects of class imbalance. 
Additionally, to tackle the issue of noisy pseudo-labels~\cite{Noisy_Pseudo, NoisyPL1} in SSOD, a two-stage filtering technique~\cite{filtering2, filtering1} is suggested to produce accurate pseudo-labels.

\subsection{SED} 
\begin{figure}[H]
\centering
\includegraphics[width=1.00\linewidth]{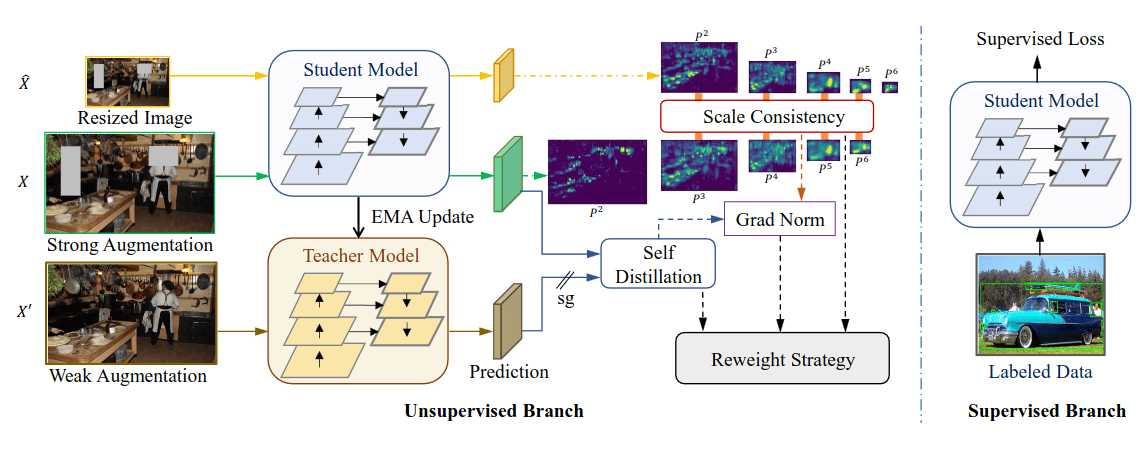}
   \caption{ \textbf{Framework of SED~\cite{SED97}} }
\label{fig:SED}
\vspace{-10pt}
\end{figure}

An innovative method called Scale-Equivalent Distillation (SED)~\cite{SED97} introduces an end-to-end knowledge distillation framework~\cite{Know_dis_SED} that is both straightforward and efficient. SED diminishes noise from erroneous negative data, enhances localization accuracy, and deals with high object size variance by enforcing scale consistency regularization~\cite{Consistency_SED}, as represented in Fig. \ref{fig:SED}, . Furthermore, a re-weighting technique~\cite{Reweight1} effectively minimizes class imbalance~\cite{imbalance_ACRST1,imbalance_ACRST2,imbalance_ACRST3,imbalance_ACRST4} by implicitly identifying potential foreground areas from unlabeled data.
\subsection{SCMT} 
\begin{figure}[H]
\centering
\includegraphics[width=0.90\linewidth]{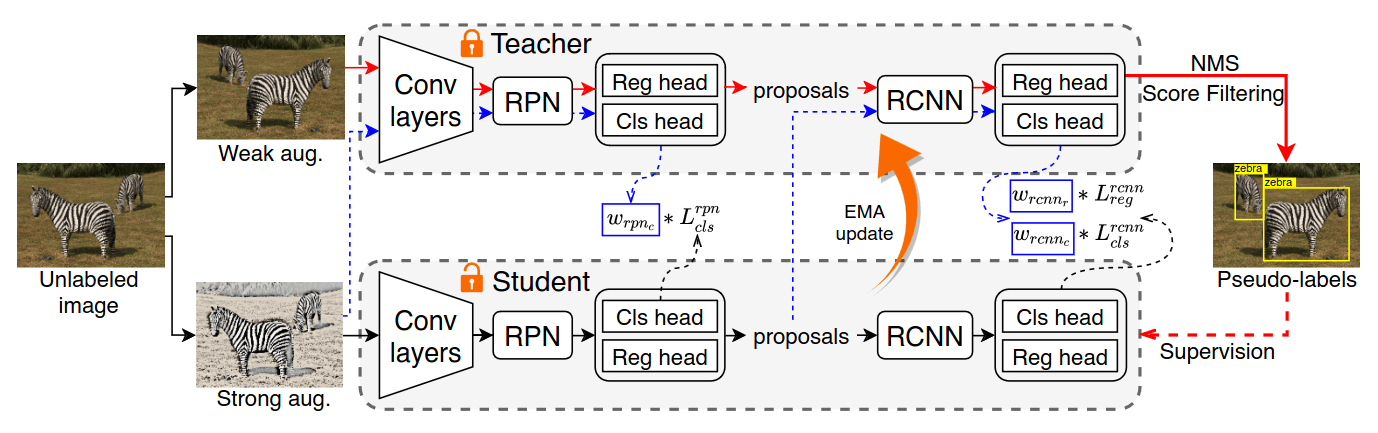}
   \caption{ \textbf{Framework of Self-Correction Mean Teacher ~\cite{SCMT22}} }
\label{fig:SCMT}
\vspace{-10pt}
\end{figure}

The objective of Self-Correction Mean Teacher(SCMT)~\cite{SCMT22} is to reduce the negative impact of noise present in pseudo-labels~\cite{intropseudo,intropseudo2,intropseudo3} by dynamically modifying loss weights for box candidates. Depicted in Fig. \ref{fig:SCMT}, SCMT effectively prioritizes more reliable box candidates during training by utilizing confidence scores derived from both localization accuracy~\cite{pseco_localization_2} and classification scores. This novel approach outperforms existing methods~\cite{STAC54,HumbleTeacher56,unbiased_Teacher_ICLR21}, demonstrating its potential to improve the performance of object detection models in real-world contexts.
\subsection{Omni-DETR} 
\begin{figure}[H]
\centering
\includegraphics[width=0.95\linewidth]{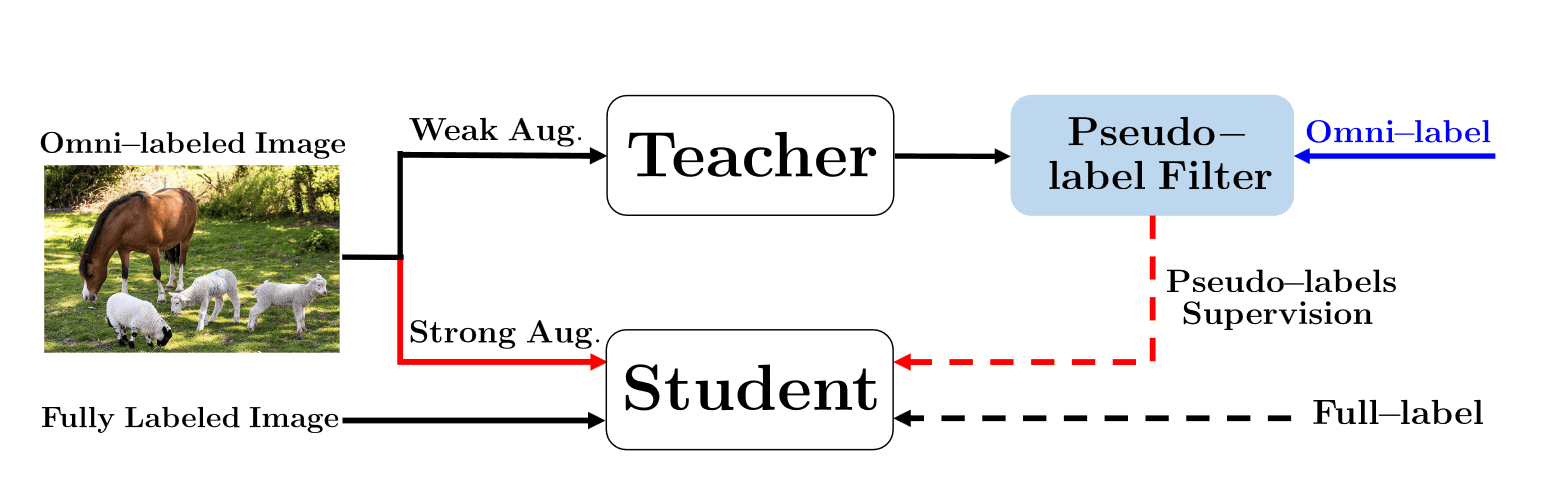}
   \caption{ \textbf{Framework of Omni-DETR~\cite{omni-DETR_cvpr22}} }
\label{fig:Omni-Detr}
\vspace{-10pt}
\end{figure}

In order to improve detection accuracy while lowering annotation costs, the Omni-DETR~\cite{omni-DETR_cvpr22}  framework is shown in Fig. \ref{fig:Omni-Detr}, incorporates a variety of weak annotations~\cite{weak_annotation}, including picture tags, item counts, and points. By integrating recent developments in end-to-end transformer-based detection architecture~\cite{Omni_EndtEnd,deformable_detr} and student-teacher-based Semi-Supervised Object Detection~\cite{STAC54,UnbiasedTeacher57}, Omni-DETR enables the use of unlabeled and poorly labeled data to produce precise pseudo labels~\cite{intropseudo,intropseudo2,intropseudo3} .

\subsection{Semi-DETR} 
\begin{figure}[H]
\centering
\includegraphics[width=0.99\linewidth]{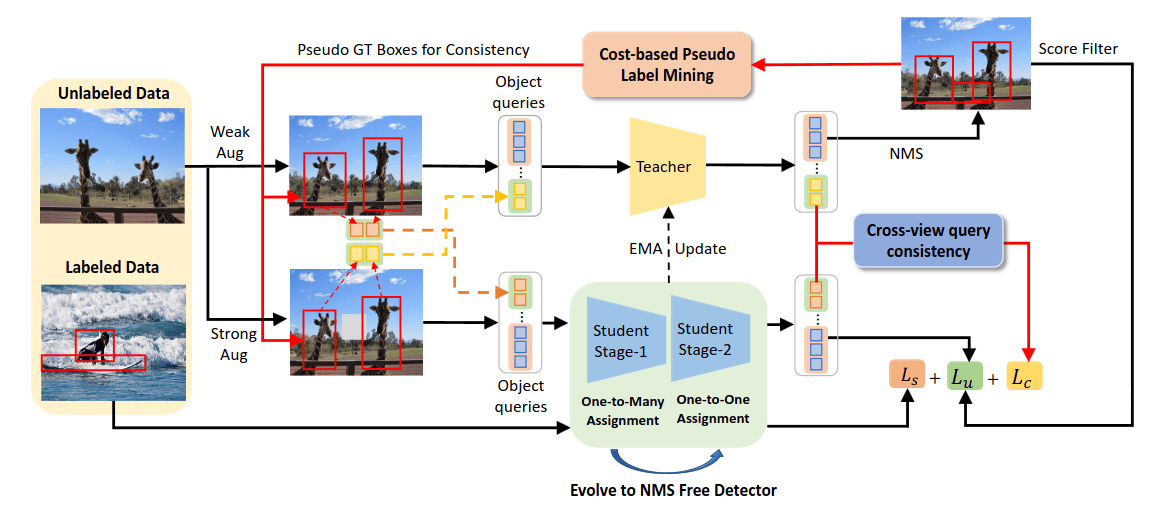}
   \caption{ \textbf{Framework of Semi-Detr~\cite{Semi-DETR_cvpr23}} }
\label{fig:Semi-Detr}
\vspace{-10pt}
\end{figure}

 Semi-DETR~\cite{Semi-DETR_cvpr23} employs a Stage-wise Hybrid Matching strategy~\cite{semihybrid} to combine one-to-one~\cite{sparse_semi_detr2} and one-to-many\cite{onetmany} assignment strategies, enhancing training efficiency and providing high-quality pseudo-labels.~\cite{intropseudo,intropseudo2,intropseudo3}. As represented in Fig. \ref{fig:Semi-Detr}, a Cross-view Query Consistency method~\cite{crossview} eliminates the need for deterministic query correspondence, facilitating the learning of semantic feature invariance. Additionally, the Cost-based Pseudo Label Mining~\cite{pseudo_labelmining} module dynamically identifies reliable pseudo boxes for consistency learning.
\subsection{Sparse Semi-DETR} 
\begin{figure}[H]
\centering
\includegraphics[width=0.99\linewidth]{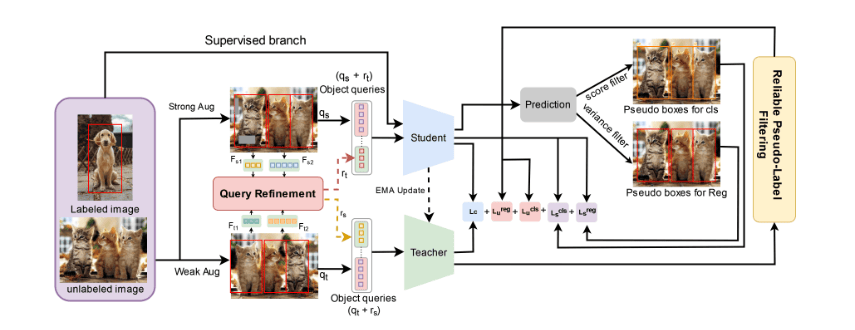}
   \caption{ \textbf{Framework of Sparse Semi-Detr~\cite{sparse_semi_detr2}} }
\label{fig:Sparse Semi-Detr}
\vspace{-10pt}
\end{figure}

Sparse Semi-DETR~\cite{sparse_semi_detr2}, an end-to-end Semi-Supervised Object Detection system based on transformers. This solution deals with problems regarding the quality of object queries in particular and resolves them. Training efficiency is slowed and model performance is gets worse by inaccurate pseudo-labels~\cite{omni-DETR_cvpr22} and redundant predictions, especially for tiny or obscured objects. To improve object query quality and greatly increase detection capabilities for tiny and partially obscured objects, Sparse Semi-DETR includes a Query Refinement Module~\cite{query1}, as illustrated in Fig. \ref{fig:Sparse Semi-Detr}. Robust pseudo-label filtering modules further improve detection accuracy and consistency by filtering only high-quality pseudo-labels~\cite{InstantTeaching59,SoftTeacher}.

\section{Loss Function}
\label{sec:loss}
\subsection{Smooth L1 Loss}
Smooth L1 loss~\cite{smoothL,smoothL2,Label94}, often used in object detection tasks, offers a gentle penalty for model errors, making it effective in scenarios with noisy or sparse data. It reduces sensitivity to outliers, contributing to more stable training and improved model performance~\cite{errors}.
\subsection{Focal Loss}
Focal loss~\cite{Focal1,Focal2} addresses class imbalance~\cite{S4OD_Imblanace,imbalance_ACRST1,imbalance_ACRST2,imbalance_ACRST3} by dynamically adjusting the importance of different examples based on their classification\cite{app_image1,app_image2} difficulty. This loss function is often integrated into strategies focusing on leveraging unlabeled data to enhance model robustness.
\subsection{Distillation Loss}
Knowledge transfer~\cite{transfer_know} from a teacher model based on labeled data to a student model with utilization of unlabeled samples is facilitated by distillation loss~\cite{disti_loss1,HumbleTeacher56}. It is frequently incorporated into semi-supervised frameworks~\cite{intro_semimethods,inrosemisupervised,inrosemisupervised2} to enhance the capacity of smaller student models to generalize.
\subsection{KL Divergence}
Employed in semi-supervised scenarios~\cite{intro_semimethods,inrosemisupervised,inrosemisupervised2} to align predictions made on labeled and unlabeled data, KL divergence loss~\cite{KL2,KL1,HumbleTeacher56,Combating92} minimizes the difference between probability distributions. It is commonly used in strategies aiming to leverage unlabeled data to improve model consistency and performance.
\subsection{Quality Focal Loss}
Quality Focal Loss~\cite{quailt_FL,Dense-Teacher74} assigns varying weights to examples based on
their difficulty levels, prioritizing the learning from challenging
instances. This loss function is often used in strategies focusing
on maximizing the use of labeled and unlabeled insight.
\subsection{Consistency Regularization Loss}
Consistency regularization loss~\cite{SED97,CSD75} ensures consistency in predictions across different views of the same input data, enhancing model robustness and generalization in SSOD. It penalizes inconsistencies, prompting the model to learn invariant features~\cite{variant}, thereby improving performance across varied datasets.
\subsection{Jensen-Shannon Divergence}
Jensen-Shannon divergence~\cite{JensenSD1,JensenSD2} regularizes ensembles by aligning predictive distributions with ground truth labels, improving prediction consistency.
\subsection{Pseudo-Labeling Loss}
Pseudo-Labeling Loss~\cite{PseudoLoss} is a technique that facilitates semi-supervised approaches~\cite{intro_semimethods,inrosemisupervised,inrosemisupervised2} by labeling unlabeled data based on model predictions and penalizing differences between expected and actual labels. It makes use of unlabeled data to boost model performance by encouraging confident predictions on samples without labels.
\subsection{Cross-Entropy Loss}
The difference between the estimated probability distribution and actual distribution of labels is measured by the Cross-Entropy Loss~\cite{Loss_cross1,STAC54,UnbiasedTeacher57}.
By encouraging the model to reduce the gap between the ground truth and predicted probabilities, this loss increases classification accuracy.
\section{Datasets and Comparison}
\label{sec:comparison}
In the object detection, having challenging datasets is crucial for ensuring fair and accurate evaluations of different algorithms. 
\subsection{Datasets}
Microsoft created the MS-COCO (Microsoft Common Objects in COntext) dataset~\cite{coco1}, that includes a wide range of images labeled with several labeling tasks, for instance segmentation and key point recognition. Counting approximately 328,000 photos and 2.5 million classified object instances over 91 categories, MS-COCO is one of the most extensive and large-scale datasets available. Semi-supervised object detection techniques can enhance model performance and generalization by combining labeled examples from COCO with unlabeled data, which eliminates the need for laborious manual annotation work.

Originating from the PASCAL Visual Object Classes Challenge, the PASCAL VOC (Visual Object Classes) dataset~\cite{voc_pascal} comprises a varied collection of photos labeled with bounding boxes and object labels spanning many categories such as household goods, cars, and animals. Each annual release from 2005 to 2012 consists of about 11,000 images for training and validation, along with an additional 10,000 images for testing. With annotations for over 27,000 object instances across 20 categories, PASCAL VOC serves as a comprehensive benchmark for evaluating object detection algorithms.
\begin{table*}[!htb]
\begin{minipage}[t]{0.45\linewidth}
\centering
\footnotesize
\begin{adjustbox}{width=\linewidth}
 \begin{tabular}{l c c c c c }
  \toprule
    \multirow{2}{*}{Methods} &
    \multirow{2}{*}{Stages}&
    \multirow{2}{*}{Reference} & 
    \multicolumn{3}{c}{COCO-Partial} \\
      \cmidrule(lr){4-6} & &  & { 1\% } & {5\%} &  { 10\%} \\
   \toprule 
   One Teacher~\cite{OneT96} & \multirow{6}{*}{One Stage}   & - & 15.4 & 36.70 & 45.3 \\
   DSL~\cite{DSL_CVPR22} & & CVPR22  &  22.03 & 30.87 & 36.22  \\
   Dense Teacher~\cite{DenseTeacher} & & ECCV22 &  22.38 & 33.01 & 37.13\\
   Unbiased Teacher v2~\cite{UnbiasedTeacherv2_CVPR22} &  & CVPR22&  22.71 & 30.08  & 32.61 \\
   S4OD~\cite{S4OD70} & & - & 23.70 & 32.30 & 35.00  \\
   Consistent-Teacher~\cite{Consistent-Teacher72} & &
   CVPR23 &  25.30 & 36.10 & 40.00 \\   
    \midrule 
    Rethinking pse~\cite{Rethinking98} & \multirow{17}{*}{Two Stage}  & AAAI22 & 9.02 & 28.40 & 32.23 \\
    CSD~\cite{CSD75} &  & ICML23& 10.51 & 18.63 & 22.46\\
    STAC~\cite{STAC54} &   & - & 13.97& 24.38& 28.64\\
    Humble Teacher~\cite{HumbleTeachers_CVPR21} &  & CVPR22 & 16.96 & 27.70 & 31.61 \\
    Instant-Teaching~\cite{Instant-Teaching_CVPR21} & & CVPR21  & 18.05 & 26.75& 30.40 \\
    ISMT~\cite{ISMT94} &   & CVPR21 & 18.41 & 26.37 & 30.53\\
    Combating Noise~\cite{Combating92} &   & - & 18.41 & 28.96 & 32.43\\
    Soft Teacher~\cite{SoftTeacher}& & ICCV21 & 20.46 & 30.74 & 34.04 \\
    Unbiased Teacher ~\cite{UnbiasedTeacher57} &  & ICLR21 & 20.75 & 28.27 & 31.50\\
    DTG-SSOD~\cite{DTG95} &   & - & 21.27 & 31.90 & 35.92 \\
    MUM~\cite{Mum95} &   & CVPR22 &  21.88 & 28.52 & 31.87 \\
    Active Teacher~\cite{Active100} &   &  CVPR22 & 22.20& 30.07 & 32.58 \\
    PseCo~\cite{Pseco99} & & ECCV22 & 22.43 & 32.50 & 36.06\\
    CrossRectify~\cite{Rectify93} &   & CVPR22 & 22.50 & 32.80 & 36.30\\
    Label Match~\cite{Label94} &   & CVPR22 & 25.81 & 32.70 & 35.49\\
    ACRST~\cite{ACRST73} &   & - & 26.07 & 31.35 & 34.92\\  
    SED~\cite{SED97} &   & CVPR22 & - & 29.01 & 34.02 \\
    SCMT~\cite{SCMT22} &   & IJCAI22 & 23.09 & 32.14 & 35.42 \\

    \midrule
    Omni-DETR~\cite{omni-DETR_cvpr22} &\multirow{3}{*}{End to End}  & CVPR22 & 27.60 & 37.70 & 41.30 \\
    Semi-DETR~\cite{Semi-DETR_cvpr23} & & CVPR23 & 30.50 & 40.10 & 43.5  \\
    Sparse Semi-DETR~\cite{sparse_semi_detr2} & & CVPR24 &  30.90 &40.80 & 44.30 \\
    \bottomrule
  \end{tabular}
  
  \end{adjustbox}
 \caption{\textbf{Object Detection Performance on COCO-Partial Dataset.} Comparison of object detection methods across different stages on the COCO-Partial dataset.}
 \label{tab:results_coco_partial}
\end{minipage}%
\hfill
\begin{minipage}[t]{0.45\linewidth}
\centering
\footnotesize
\begin{adjustbox}{width=\linewidth}
 \begin{tabular}{l c c c c c }
  \toprule
    \multirow{2}{*}{Methods} &
    \multirow{2}{*}{Stages}&
    \multirow{2}{*}{Reference} & 
    \multicolumn{3}{c}{PASCAL-VOC} \\
      \cmidrule(lr){4-6} & &  & { $AP_{50}$ } & {$mAP$} &  { $AP_{75}$} \\
   \toprule 
    S4OD~\cite{S4OD70} & \multirow{6}{*}{One Stage} & - &  50.1 & - &34.0  \\
    Dense Teacher~\cite{DenseTeacher} & & ECCV22 &  79.89 & 55.87& -\\
    DSL~\cite{DSL_CVPR22} & & CVPR22  &  80.7 & 56.8& - \\
    Consistent-Teacher~\cite{Consistent-Teacher72} & & CVPR23 &  81.00 & 59.00 & -\\   
    Unbiased Teacher v2~\cite{UnbiasedTeacherv2_CVPR22} &  & CVPR22&  81.29 & 56.87 & -\\
    One Teacher~\cite{OneT96} &   & - & 76.1 & -&- \\
 
    \midrule 
    Soft Teacher~\cite{SoftTeacher}& \multirow{17}{*}{Two Stage} & ICCV21 & 20.46 & 30.74 & 34.04 \\
    Combating Noise~\cite{Combating92} &   & - & 43.2 & 62.0 & 47.5\\
    DTG-SSOD~\cite{DTG95} &   & - & 56.4&-& 38.8 \\
    PseCo~\cite{Pseco99} & & ECCV22 & 57.2 &-&39.2\\
    CSD~\cite{CSD75} & & ICML23& 74.70 & - & -\\
    ISMT~\cite{ISMT94} &   & CVPR21 & 77.23 & 46.23& -\\
    Unbiased Teacher ~\cite{UnbiasedTeacher57} &  & ICLR21 & 77.37 & 48.69&-\\
    STAC~\cite{STAC54} &   & - & 77.45 & 44.64& - \\
    ACRST~\cite{ACRST73} &   & - & 78.16 &50.1&-\\
    MUM~\cite{Mum95} &   & CVPR22 &  78.94 & 50.22 & -\\
    Rethinking pse~\cite{Rethinking98} &   & AAAI22 & 79.0 & 54.60 &59.4 \\
    Instant-Teaching~\cite{Instant-Teaching_CVPR21} & & CVPR21  & 79.20&50.00 & 54.00\\
    Humble Teacher~\cite{HumbleTeachers_CVPR21} &  & CVPR22 & 80.94 & 53.04 & - \\
    CrossRectify~\cite{Rectify93} &   & CVPR22 & 82.34 &- &- \\
    Label Match~\cite{Label94} &   & CVPR22 & 85.48 & 55.11& -\\
    SED~\cite{SED97} &   & CVPR22 & 80.60 & - & - \\
     \midrule
    Semi-DETR~\cite{Semi-DETR_cvpr23} & \multirow{3}{*}{End to End} & CVPR23 & 86.10 &65.2 & -  \\
    Sparse Semi-DETR~\cite{sparse_semi_detr2} & & CVPR24 &  86.30 & 65.51 & - \\

    \bottomrule
  \end{tabular}

  \end{adjustbox}
 \caption{\textbf{Object Detection Performance on PASCAL-VOC Dataset.} Comparison of object detection methods across different stages on the PASCAL-VOC dataset.}
 \label{tab:results_pascal_voc}
\end{minipage}
\end{table*}

\begin{figure*}[h]
\centering
\includegraphics[width=0.9\textwidth]{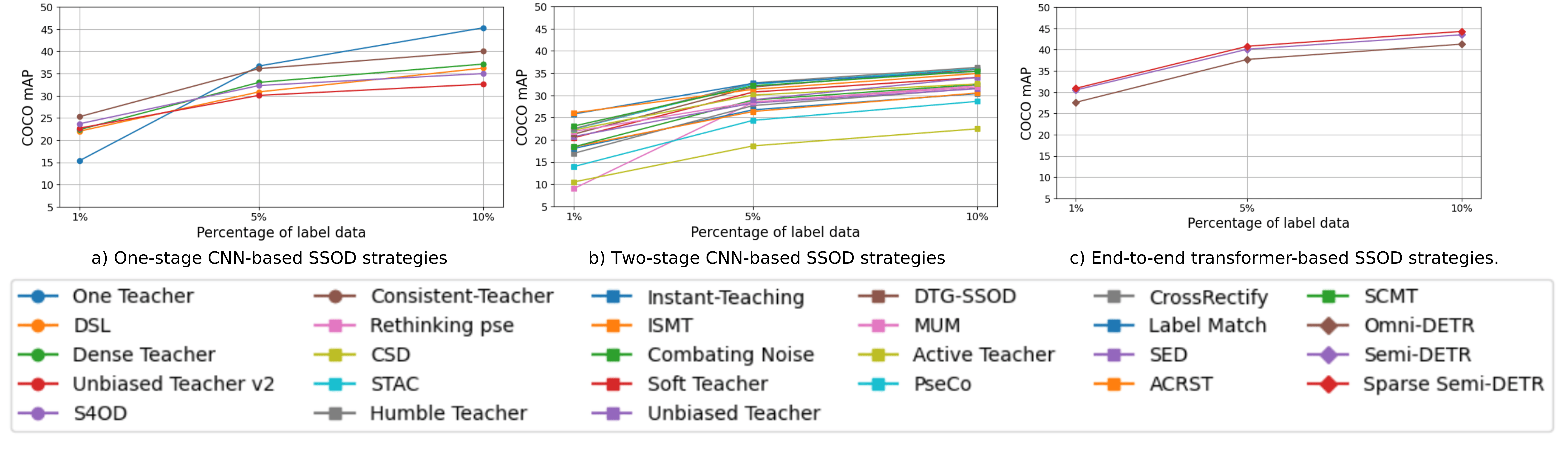}
\caption{Comparison of CNN-based and transformer-based SSOD strategies on COCO dataset. (a) Performance comparison of one-stage CNN-based SSOD Strategies. (b) Performance comparison of two-stage CNN-based SSOD Strategies. (c) Performance comparison of end to end transformer-based SSOD Strategies.}\label{fig:graphscoco}
\end{figure*}
\subsection{Comparison}
The performance of object detection methods has been extensively evaluated on benchmark datasets such as COCO and PASCAL. These evaluations show the progress and effectiveness of both one-stage and two-stage detection approaches, as well as end-to-end methods, in improving detection accuracy over various training epochs.\par
Table \ref{tab:results_coco_partial} offers the performance comparison of various methods on COCO dataset~\cite{coco1}. One stage methods, including One Teacher~\cite{OneT96}, DSL~\cite{Denselearning60}, Dense Teacher~\cite{Dense-Teacher74}, demonstrate incremental improvements with increasing training epochs. Two stage methods, such as Rethinking pse~\cite{Rethinking98}, STAC~\cite{STAC54}, and Combating Noise~\cite{Combating92}, exhibit consistent enhancement in performance metrics over epochs. Notably, DETR-based models like Omni-DETR~\cite{omni-DETR_cvpr22} and Semi-DETR~\cite{Semi-DETR_cvpr23} showcase significant performance gains, highlighting the effectiveness of Semi-Supervised Object Detection strategies, as shown in Fig. \ref{fig:graphscoco}.

\begin{figure}[H]
\centering
\includegraphics[width=\linewidth]{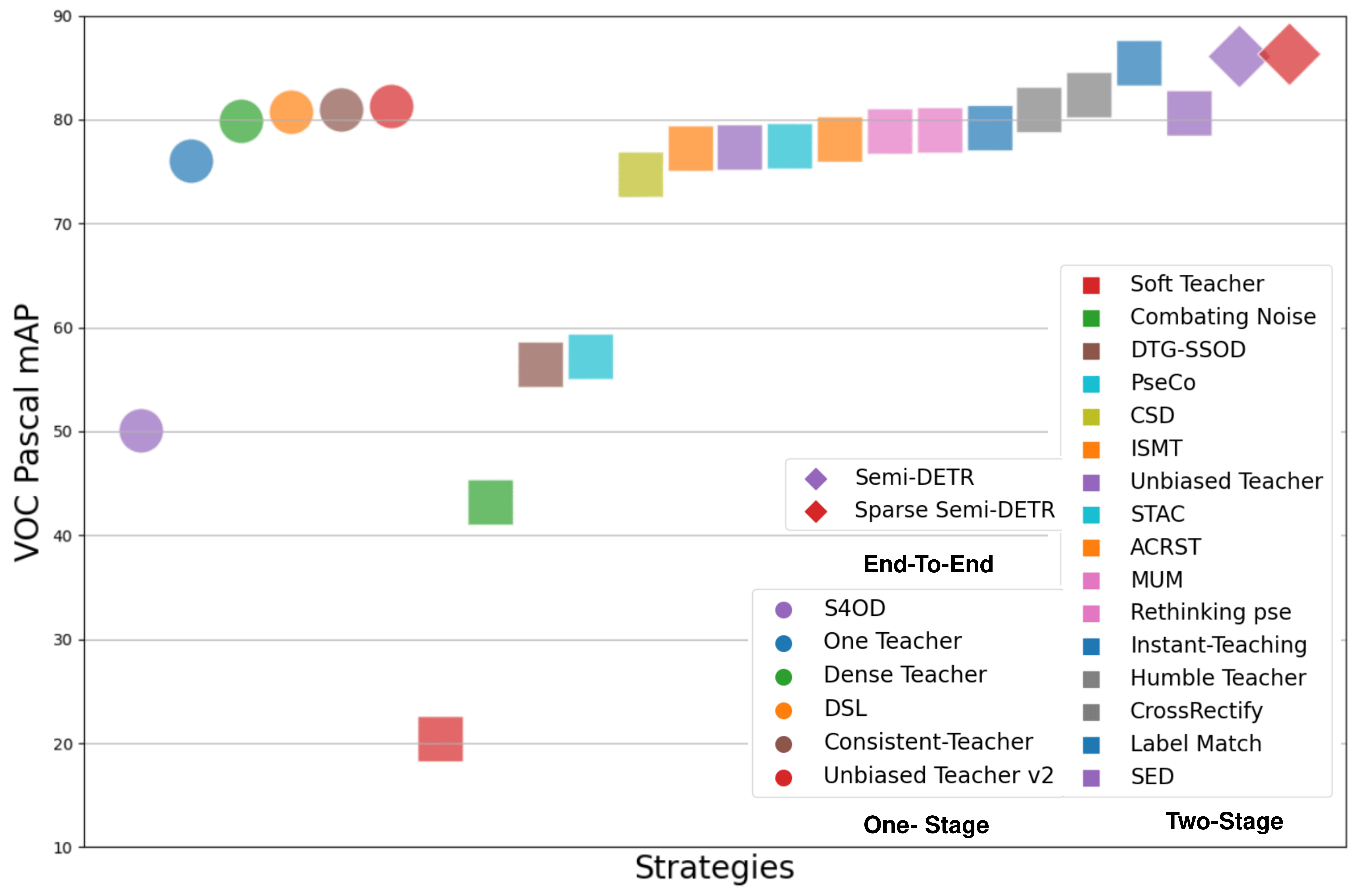}
\caption{Comparison of CNN-based (one-stage, two-stage) and transformer-based (end-to-end) SSOD strategies on VOC pascal dataset.}\label{fig:graphspascal}
\end{figure}
Table \ref{tab:results_pascal_voc} shows the performance metrics of various object detection methods across different stages on the PASCAL dataset~\cite{voc_pascal}. In the One stage, methods like S4OD~\cite{S4OD70}, Dense Teacher~\cite{Dense-Teacher74}, DSL~\cite{Denselearning60} exhibit competitive performance in terms of AP50, AP50.95, and AP75 scores. Two-stage methods like Soft Teacher~\cite{SoftTeacher58}, Combating Noise~\cite{Combating92}, and Instant-Teaching~\cite{InstantTeaching59} display significant variations in performance across different metrics. Finally, end-to-end methods like Semi-DETR~\cite{Semi-DETR_cvpr23} and Sparse Semi-DETR~\cite{sparse_semi_detr2} showcase significant performance, indicating the efficacy of SSOD approaches, as illustrated in \ref{fig:graphspascal}.

\begin{table*}
\tiny
\begin{center}
\caption{A brief description of Advantages and limitations of Semi Supervised Strategies}\label{tab:adv-limit}
\renewcommand{\arraystretch}{0.6} 
\begin{tabular*}{\textwidth}
{@{\extracolsep{\fill}}p{1.9cm}p{6.5cm}p{6.5cm}@{\extracolsep{\fill}}}
\toprule
\textbf{Methods} & 
\textbf{Advantages} & 
\textbf{Limitations} \\
\toprule
Stac \cite{STAC54} &  Improves detection performance with minimal complexity.& Low performance with frameworks employing intense hard negative mining, leading to over dependence on noisy pseudo-labels. \\
\midrule
Humble Teacher \cite{HumbleTeacher56} &  Improves performance significantly with dynamic teacher model updates and soft pseudo-labels. & More computational resources due to the dynamic updating of the teacher model and the ensemble of numerous teacher models, potentially increasing training time and complexity.  \\
\midrule
Instant Teaching \cite{InstantTeaching59} & Improving model learning with extended weak-strong data augmentation as well as instant pseudo labeling . &  Dependency on Extensive weak-strong data augmentation and instant pseudo labeling introduce computational overhead, increase training complexity and time.\\
\midrule
Soft Teacher \cite{SoftTeacher58}  & Enhances detector performance and pseudo label quality simultaneously. & Depending on extensive data augmentation and the soft teacher approach potentially increase training complexity and computational overhead. \\
\midrule
Unbiased Teacher \cite{unbiased_Teacher_ICLR21}& Effectively mitigates pseudo-labeling bias and overfitting in Semi-Supervised Object Detection.& Relies on the balance between the student and teacher models, which require careful tuning and additional computational resources.\\
\midrule
ACRST \cite{ACRST73} & Improves performance by addressing class imbalance. & Effectiveness relies on the precision of pseudo-labels, which are impacted by noise due to the complexity of detection tasks, requiring robust filtering mechanism.\\
\midrule
Combating Noise\cite{Combating92} & Effectively combating noise associated with pseudo labels enhances the robustness of the SSOD Tasks.& Dependence on accurately quantifying region uncertainty is challenging in complex scenes or datasets with diverse object characteristics.\\
\midrule
MUM \cite{Mum95}& Effectively augments data for Semi-Supervised Object Detection, enhancing model robustness without significant computational overhead. & Encounter difficulties in accurately locating object boundaries due to the mixing process, potentially affecting localization precision. \\
\midrule
 ISTM\cite{ISTM55} & Effectively leveraging ensemble learning to enhance the usefulness of pseudo labels and stabilize Semi-Supervised Object Detection training. & Introduce additional computational complexity due to the ensemble approach and the use of multiple ROI heads, potentially increasing training time and resource requirements. \\
\midrule
 Cross Rectify\cite{Rectify93} & Enhances pseudo label quality and detection performance by rectifying misclassified bounding boxes using detector disagreements. & Simultaneous training of two detectors increase computational overhead, potentially prolonging training time and resource usage.\\
\midrule
SED \cite{SED97}  & Improves Semi-Supervised Object Detection by enforcing scale consistency and self-distillation. & Reliance on the IoU threshold criterion, which could not be optimal for all detectors and situations, and its limited benefits from multi-scale testing\\
\midrule
Label Match\cite{Label94}  & Improves Semi-Supervised Object Detection by addressing label mismatch through distribution-level and instance-level methods. & Assumes Both unlabeled as well as labeled data have the same distribution, potentially restricting its applicability in diverse scenarios.\\
\midrule
DTG-SSOD\cite{DTG95}  & Leverages Dense Teacher Guidance for more accurate supervision, enhancing Semi-Supervised Object Detection performance. & Implementation complexity, especially with Inverse NMS Clustering and Rank Matching, increase computational resources during training.\\
\midrule
Rethinking Pse \cite{Rethinking98}   & Certainty-aware pseudo labels improve performance by addressing localization precision and class imbalance issues &  Implementing certainty-aware pseudo labeling add additional computational complexity during training. \\
\midrule
CSD \cite{CSD75} & Leverages consistency constraints for both classification and localization, enhancing object detection performance using unlabeled data.& It shows less performance improvement in two-stage detectors compared to single-stage detectors.\\
\midrule
PseCo \cite{Pseco99} & Enhances SSOD by integrating object detection attributes into pseudo labeling along with consistency training, leading to superior performance and faster convergence. & Its potential struggle with generalization across diverse datasets due to variability in pseudo-label quality.

\\
\midrule
Active Teacher \cite{Active100} &  Maximizes limited label information through active sampling, enhancing pseudo-label quality and improving SSOD performance. & Require more training steps compared to other methods, potentially increasing computational overhead.
\\
\midrule
One Teacher \cite{OneT96} & Improves SSOD on YOLOv5, tackling issues like low-quality pseudo-labeling. & Lowering the threshold for pseudo-labeling due to noisy pseudo-labeling in one-stage detection makes it difficult to maximize the effectiveness of one-stage teacher-student learning. \\
\midrule
Dense Teacher \cite{Dense-Teacher74} & Simplifies the SSOD pipeline by using Dense Pseudo-Labels, improving efficiency and performance. & Contain high-level noise, potentially impacting detection performance if not properly addressed.\\
\midrule
Unbiased Teacher v2\cite{unbiased_Teacher_ICLR21} & Expands the applicability of SSOD to anchor-free detectors, improving performance across various benchmarks.& Challenges remain in scaling the method to large datasets, integrating localization uncertainty estimation for boundary prediction with the relative thresholding mechanism, and addressing domain shift issues.\\
\midrule
S4OD\cite{S4OD70} & Dynamically adjusts pseudo-label selection to balance quality and quantity, enhancing single-stage detector performance & DSAT strategy's increased time cost is due to F1-score computation, and using the CPU version of NMS for uncertainty computation slows down training. \\
\midrule
Consistent-Teacher\cite{Consistent-Teacher72} & Improves SSOD performance by addressing inconsistent pseudo-targets with feature alignment, adaptive anchor assignment, and dynamic threshold adjustment. &Performance is validated mainly on single-stage detectors, with effectiveness on stage-two detectors and DETR-based models yet to be confirmed.\\
\midrule
Omni-DETR \cite{omni-DETR_cvpr22} & utilize diverse weak annotations to enhance performance and annotation efficiency.&  Effectiveness on larger datasets is uncertain, and its simplified annotation process could raise concerns about potential misuse.
\\
\midrule
Semi-DETR \cite{Semi-DETR_cvpr23} & Combines Cross-view query consistency and stage-wise hybrid matching to improve training efficiency. & Encounter challenges due to the absence of deterministic connection between the predictions and the input queries.\\
\midrule
Sparse Semi-DETR \cite{sparse_semi_detr2} & Introduces a Query Refinement Module to improve object query functionality, enhancing detection performance for small and obscured objects.& Require additional computational resources due to the integration of novel modules, potentially increasing training time and complexity.\\

\bottomrule
\end{tabular*}
\end{center}
\end{table*} 
\section{Open Challenges \& Future Directions}
\label{sec:Discussion}
Semi-Supervised Object Detection (SSOD) has shown remarkable progress, transitioning from traditional Convolutional Neural Networks (CNNs) to advanced Transformer-based models. This survey presents a comprehensive overview of SSOD methods, highlighting their advantages and addressing the challenges they face. The area of Semi-Supervised Object Detection (SSOD) has drawn proposals for numerous methods to leverage unlabeled data and enhance detection performance. These methods have certain benefits and drawbacks. Table \ref{tab:adv-limit} provides a detailed summary of their benefits and drawbacks. 
Despite its thorough examination of Semi-Supervised Object Detection methodologies, the survey's broad focus might result in overlooking some specific approaches or recent advancements in the field.
The scope of the survey may limit the depth of analysis for each semi-supervised approach, potentially sacrificing detailed insights into their underlying principles, advantages, and limitations.

 Given the diverse range of semi-supervised algorithms tailored for object detection tasks, a more in-depth examination of each methodology could provide a richer understanding of their efficacy and applicability across different domains. 
While these methods represent some of the most intuitive approaches to SSOD, they still have many obstacles. Looking ahead, there are some potential possibilities for future improvements.\par
\textbf{Domain Adaptation and Transfer Learning:} Enhancing the generalizability of Semi-Supervised Object Detection models requires exploring domain adaptation as well as transfer learning techniques. Adapting models trained on synthetic or labeled datasets to real-world domains with limited labeled data is essential for practical deployment.\par
\textbf{Hybrid Approaches and Model Compression:}
Investigating hybrid approaches that integrate semi-supervised object detection with transfer learning, self-supervised learning, or model compression can improve the efficiency and effectiveness of object detection systems. Novel hybrid architectures and training strategies can lead to resource-efficient and scalable solutions.


\section{Applications}
\label{sec:application}
\subsection{Image Classification}
Semi-supervised learning has significantly advanced image classification~\cite{app_image1,app_image2}, especially in fields with limited labeled data~\cite{app_image9}. In medical imaging~\cite{app_image3,app_image4}, it enables precise disease classification from X-rays and MRIs with few labeled samples. Remote sensing~\cite{app_image5,app_image6} benefits by improving land cover and environmental change classification from satellite images, aiding urban planning and disaster management. For autonomous vehicles~\cite{app_image7,app_image8}, semi-supervised learning enhances the classification of objects and pedestrians, promoting safer navigation. Techniques such as consistency regularization~\cite{Consistency_SED} and pseudo-labeling~\cite{intropseudo,intropseudo2,intropseudo3} have been crucial in refining these models, increasing their robustness and accuracy.
\subsection{Document Analysis}
Semi-Supervised Object Detection is increasingly applied in document analysis~\cite{bridging_per3,shehzadi2024hybrid6} to efficiently identify and classify elements such as text blocks, tables, and images within documents.~\cite{app_doc1,app_doc_3,app_doc_4,app_doc_5,unsupdla10}. This approach is particularly valuable in legal, financial, and academic fields, where large volumes of documents need to be processed~\cite{app_doc2,app_doc_6,bankchecksecurity11}. By leveraging both labeled and unlabeled data, semi-supervised methods~\cite{shehzadi_semi-detr_table1,shehzadi2024endtoend7,ehsan_semi8} improve the accuracy and efficiency of detecting critical information like clauses, dates, amounts, and references.~\cite{app_doc_8,app_doc_7,app_doc_9}. Techniques such as consistency regularization~\cite{Consistency_SED} and self-training~\cite{Self_Training_SSOD_CVPR20} enhance model robustness, making document analysis more automated and reliable despite limited labeled data.
\subsection{3D Object Detection}
Semi-Supervised Object Detection significantly enhances 3D detection~\cite{app_3d1,app_3d2} applications by leveraging both labeled and unlabeled data to improve accuracy and robustness. In autonomous driving~\cite{app_3d3,app_3d4}, it allows vehicles to better identify and classify objects like pedestrians and obstacles using LIDAR~\cite{app_3d5,app_3d6} and camera data~\cite{app_3d7}~\cite{app_3d8}, enhancing safety and navigation. In robotics~\cite{app_3d9,app_3d10}, it aids in precise object manipulation and obstacle avoidance. Additionally, in augmented and virtual reality~\cite{app_3d11}, it enables more immersive experiences by accurately integrating digital elements with the real world. These advancements make Semi-Supervised Object Detection a crucial technology for various 3D detection tasks.
\subsection{Network traffic Classification}
Semi-Supervised Object Detection is also effectively applied in network traffic classification~\cite{app_NTC,app_NTC2,app_NTC3} , where it helps identify and categorize various types of network traffic~\cite{wajahatCC9} with limited labeled data. By leveraging both labeled and unlabeled traffic data, these models can more accurately detect patterns, anomalies, and potential security threats in network activity. This approach enhances the ability to manage and secure networks, improving the detection of malicious activities~\cite{app_NTC_mal} such as intrusions~\cite{app_NTC_Intrusion} and data breaches while ensuring efficient network performance. Semi-supervised learning thus plays a crucial role in maintaining robust and secure network infrastructures.
\subsection{Speech Recognition}
In speech recognition~\cite{app_speech,app_speech2,app_speech3,app_speech4} , SSOD aids in identifying and classifying speech patterns and phonetic elements within audio data, even with limited labeled samples. By leveraging both labeled and unlabeled speech data, these models can better discern speech signals from background noise and accurately transcribe spoken words into text. This approach enhances the performance and efficiency of speech recognition systems~\cite{app_speech_sys,app_speech_sys2}, enabling more accurate and reliable transcription in various applications~\cite{app_speech_app2,app_speech_app3} such as virtual assistants, dictation software, and voice-controlled devices. Additionally, SSOD techniques contribute to the scalability and adaptability of speech recognition systems, allowing them to handle diverse linguistic contexts and acoustic environments with improved accuracy.
\subsection{Drug Discovery and Bioinformatics}
In drug discovery~\cite{app_drug,app_drug2} and bioinformatics~\cite{app_bio,app_bio2}, Semi-Supervised Object Detection (SSOD) optimizes the identification and classification of molecular structures~\cite{app_bio_mol,app_bio_mol2} and biological entities~\cite{app_bio_entity,app_bio_entity2}. By leveraging both labeled and unlabeled data, SSOD accelerates the screening process for potential drug candidates and aids in target validation. This approach enhances efficiency in molecular analysis, enabling deeper insights into disease mechanisms and facilitating the development of precision medicine strategies for improved patient outcomes.

\section{Conclusion}
\label{sec:conclusion}
Semi-Supervised Object Detection (SSOD) has drawn numerous methods for leveraging unlabeled data to enhance detection performance, evolving from traditional Convolutional Neural Networks (CNNs) to modern Transformer-based models. We have analyzed the performance of these strategies on benchmark datasets such as COCO and VOC, highlighting significant improvements in detection accuracy and efficiency. This review provides a comprehensive overview of SSOD methods, highlighting their advantages, while addressing common challenges. The transition to Transformer-based models represents a substantial leap in SSOD capabilities, providing new insights and approaches for tackling complex detection scenarios. This survey aims to inspire ongoing research and innovation in SSOD, encouraging researchers to develop and refine strategies that will further contribute to the field of computer vision and its applications.

\ifCLASSOPTIONcompsoc
  \section*{Acknowledgments}
\else
  \section*{Acknowledgment}
\fi
 The work has been partially funded by the European project AIRISE under Grant Agreement ID 101092312.
\ifCLASSOPTIONcaptionsoff
  \newpage
\fi
\newpage



\bibliographystyle{IEEEtran}
\bibliography{sn-bibliography}
\end{document}